\pdfoutput=1
\documentclass[11pt]{article}

\usepackage[preprint]{acl}

\usepackage{times,latexsym}
\usepackage[T1]{fontenc}
\usepackage[utf8]{inputenc}
\usepackage{microtype,inconsolata}

\usepackage{graphicx}
\usepackage{booktabs,multirow}
\usepackage{subcaption}

\usepackage{amsmath,amssymb}

\usepackage{xcolor}

\usepackage{enumitem}

\usepackage{cleveref}

\usepackage{adjustbox}

\newcommand{\eg}{\textit{e.g.}}

\newcommand{\maxprob}{\textsc{MaxProb}}
\newcommand{\minprob}{\textsc{MinProb}}
\newcommand{\FSS}{\textsc{FSS}}
\newcommand{\LIS}{\textsc{LIS}}
\newcommand{\TempCore}{\textsc{TempCore}}

\title{\TempCore{}: Are Video QA Benchmarks Temporally Grounded?}

\author{
Hyunjong Ok \quad Jaeho Lee \\
POSTECH \\ 
\texttt{hyunjong.ok@gmail.com, jaeho.lee@postech.ac.kr}
}
\begin{document}
\maketitle

\begin{abstract}
Vision-language models (VLMs) can ingest only a limited number of video frames,
making frame selection a practical necessity.
But do current Video QA benchmarks genuinely require
temporal frame selection,
or can most questions be answered regardless of which frames are shown?
We introduce \textbf{Frame Selection Sensitivity} (\FSS{}),
a per-sample diagnostic that measures how much VLM accuracy
changes when the most relevant frames are replaced with the least relevant ones.
Across six benchmarks and eight VLMs,
we find that a large majority of samples are frame-agnostic:
only a minority are genuinely sensitive to frame choice.
Combining \FSS{} with a Language Independence Score (\LIS{})
reveals that merely 5.5--31\% of samples are \textit{Temporally Sensitive}.
We construct \TempCore{}, compact evaluation subsets
that isolate these temporal samples from existing benchmarks,
and will release code and per-sample annotations upon publication.
\end{abstract}

\section{Introduction} \label{sec:intro}

Large language models (LLMs) have transformed natural language processing,
and their extension to vision-language models (VLMs) has brought
comparable advances to image
understanding \citep{li2024llava,zhu2025internvl3,qwen25vl_arxiv2025}. These developments have spurred growing interest in applying VLMs to
\emph{video} understanding, where temporal dynamics---event ordering,
causal transitions, and changes over time---are fundamental.

Extending VLMs to video processing, however, is fundamentally constrained by input capacity:
videos may contain thousands of frames, but current models can process only around 32 at a time~\citep{lin2024video,tang2025adaptive}.
The standard approach is to rank frames by relevance
using encoders such as SigLIP \citep{zhai2023sigmoid,tschannen2025siglip}
or CLIP \citep{clip_icml2021},
and retain the top-$K$ frames \citep{tang2025adaptive,himakunthala2023let,shen2024longvu}. Accordingly, a growing line of work explores more sophisticated selection
strategies---including combinatorial ranking,
spatiotemporal saliency,
hierarchical clustering,
and generative index prediction \citep{frame_voyager_iclr2025,mllm_frame_selection_cvpr2025,videotree_cvpr2025,gens_acl2025}---all
based on the premise that selecting the right frames is critical.

While these methods report gains on standard benchmarks, a closer look
complicates the picture. On some benchmarks, simple uniform sampling
performs comparably to learned
selection~\citep{frame_sampling_strategies_arxiv2025},
and a growing literature suggests that VLMs are largely
``temporally blind,'' exhibiting limited sensitivity to event ordering
and duration~\citep{time_blindness_arxiv2025,consistency_videollm_cvpr2025,temporalbench_neurips2024}.
If temporally insensitive models can still achieve high scores,
the benchmarks themselves may not genuinely require temporal reasoning---and
if most questions can be answered from a single frame or from text alone,
then frame selection is irrelevant regardless of the method.
This tension raises two intertwined questions:
\begin{enumerate}[topsep=2pt,parsep=0pt,itemsep=2pt]
  \item \emph{Does key-frame selection actually improve Video QA performance?}
  \item \emph{Do current benchmarks genuinely require temporal reasoning?}
\end{enumerate}

To address these questions, we examine whether encoder-based frame
relevance translates into meaningful differences in downstream VLM
accuracy.
Concretely, we score every video frame with a SigLIP
encoder and compare VLM accuracy when the model receives the most
relevant (\maxprob{}) versus least relevant (\minprob{}) frames.
This yields \textbf{Frame Selection Sensitivity} (\FSS{}), a per-sample
diagnostic that partitions each benchmark sample into \emph{Key}
(temporally sensitive), \emph{Non-key} (frame-agnostic), or
\emph{Anti-key} (hurt by ``relevant'' frames).
Applying \FSS{} to six benchmarks
with 8 VLMs ranging from 256M to 8B parameters---the
largest systematic study of frame selection to date---we answer:

\paragraph{Answer to Q1: Frame selection works, but only on long videos.}
On long-video benchmarks, \maxprob{} consistently outperforms \minprob{},
while on short-video benchmarks, the effect is near zero.
Our approach uses a general-purpose image-text encoder (SigLIP)
in question-answer (QA-mode) scoring, which leverages full question-answer
context for more accurate temporal classification;
encoder improvements---such as question-aware models---could
narrow this gap further.
The large oracle headroom confirms that
substantial temporal signal exists in the video but remains unexploited
by current selection methods.

\paragraph{Answer to Q2: Most existing benchmarks are not temporally grounded.}
Only 6--36\% of samples are Key;
the vast majority are Non-key or Anti-key.
A complementary \textbf{Language Independence Score} (\LIS{})
reveals that only 5.5--31.0\% of samples are \emph{Temporally Sensitive}---both
vision-dependent and frame-sensitive (\S\ref{ssec:two_stage}).
No-frame baselines reach up to 50\% accuracy on language priors alone,
and aggregate scores are dominated by Non-key samples,
allowing models with strong language priors to appear competitive
even when they fail on frame-sensitive temporal questions.

Based on this analysis, we construct \TempCore{}, temporally grounded
evaluation subsets that retain only these samples
(\S\ref{sec:tempcore}), providing a focused measure of frame-selection
sensitivity.

\paragraph{Contributions.}
\begin{enumerate}[topsep=2pt,parsep=0pt,itemsep=2pt]
  \item We propose \FSS{}, a per-sample diagnostic
    that quantifies temporal sensitivity,
    and conduct a systematic frame selection study
    (\S\ref{sec:method}, \S\ref{sec:results}).
  \item We introduce a two-dimensional \FSS{}$\times$\LIS{} taxonomy
    that reveals only 5.5--31.0\% of samples are \emph{Temporally Sensitive}
    (\S\ref{sec:analysis}, \S\ref{sec:tempcore}).
  \item We construct \TempCore{}, temporally grounded evaluation subsets
    extracted from existing benchmarks without manual re-annotation
    (\S\ref{sec:tempcore}).
\end{enumerate}

\section{Related Work}
\label{sec:related}

\paragraph{Large language models and multimodal LLMs.}
The expansion of large language models (LLMs) has enabled
strong performance across various general-purpose language
tasks~\citep{achiam2023gpt, grattafiori2024llama, yang2024qwen2, team2025gemma}.
Building upon this foundation, research has extended such capabilities
to other modalities by leveraging the knowledge of
LLMs~\citep{alayrac2022flamingo, liu2023visual, chu2024qwen2, bai2025qwen2, zhu2025internvl3}.
Video MLLMs, in particular, have achieved strong performance
in video question answering, captioning, and
summarization~\citep{zhang2023video, lin2024video, zhulanguagebind, maaz2024video, shen2024longvu, zhang2025videollama}.

\paragraph{Key frame selection.}
Key frame selection reduces the computational cost of long-video understanding
by identifying the most relevant frames for a given query.
Encoder-based methods score individual frames via vision-language similarity:
\citet{tang2025adaptive} select frames that are both query-relevant
and visually diverse to avoid redundancy,
\citet{frame_voyager_iclr2025} rank frame subsets by how well
they complement each other rather than scoring frames independently,
and \citet{mllm_frame_selection_cvpr2025} leverage multimodal LLM prompting
to derive spatial importance and temporal selection signals for frame ranking.
A second line of work explores structured selection strategies:
\citet{videotree_cvpr2025} build a hierarchical tree of visual segments
that enables query-adaptive retrieval from long videos,
\citet{ffs_cvpr2025} learn to dynamically adjust how many frames
to select per query instead of using a fixed budget,
and \citet{mdp3_iccv2025} optimize frame orderings using
a list-wise ranking objective that considers the full sequence jointly.
These methods \emph{improve} selection;
we \emph{diagnose} when selection matters
and construct evaluation subsets that isolate temporal ability.

\paragraph{Temporal diagnostics \& benchmarks.}
Several studies document that VLMs struggle with temporal reasoning:
near-zero accuracy on ordering tasks~\citep{time_blindness_arxiv2025},
inconsistency across equivalent temporal queries~\citep{consistency_videollm_cvpr2025},
and failures in fine-grained temporal understanding~\citep{temporalbench_neurips2024}.
Complementary work shows that many Video QA questions
can be solved from a single frame or without temporal reasoning:
\citet{tvbench_bmvc2025} show that most MVBench tasks
can be solved without temporal reasoning, with single-frame and text-only
baselines remaining competitive;
\citet{lei2023revealing} demonstrate similar single-frame biases
across multiple video-language benchmarks.
On the diagnostic side,
counterfactual interventions~\citep{counterfactual_vqa_cvpr2021},
text-bias reduction~\citep{text_bias_mcqa_arxiv2026},
confidence intervals~\citep{bootstrap_ci_arxiv2025},
and sampling strategy comparisons~\citep{frame_sampling_strategies_arxiv2025}
have been proposed to isolate specific biases.
Dedicated temporal benchmarks take complementary approaches.
TempCompass~\citep{liu2024tempcompass} constructs synthetic test cases
by applying video-level manipulations---reversal, spatial concatenation,
and temporal concatenation---to probe whether models truly perceive
temporal order rather than relying on static appearance.
TemporalBench~\citep{temporalbench_neurips2024} instead enlists expert
annotators to label fine-grained temporal events within each clip,
producing densely annotated samples that test ordering, duration,
and causal sequencing.
\TempCore{} differs on two fronts:
(i)~it provides a fully automated, model-based diagnostic pipeline
that requires no manual annotation or synthetic video generation, and
(ii)~it draws from six widely adopted benchmarks covering egocentric, perceptual, and causal-reasoning domains, yielding a compact yet diverse evaluation set.

To our knowledge, \FSS{}--\LIS{} is the first model-side diagnostic that
assigns per-sample temporal-sensitivity labels to existing Video QA benchmarks
without additional annotation. Rather than asking how to select frames more
effectively, as prior frame-selection methods do, we use the
\maxprob{}/\minprob{} contrast together with a no-frame language-prior axis
to audit whether benchmark samples actually depend on temporal visual evidence.
Dedicated temporal benchmarks instead construct new test cases or annotations;
our approach identifies temporally grounded samples directly within benchmarks
already in widespread use.

\section{Diagnostic Framework}
\label{sec:method}

\subsection{Frame Scoring}

Given a video $V = (f_1, f_2, \ldots, f_L)$ of $L$ frames
uniformly extracted at 10 FPS,
and a text query $t$ (question, or concatenation with answer),
we compute a per-frame relevance score using a pretrained
SigLIP vision-language encoder~\citep{zhai2023sigmoid,tschannen2025siglip}:
\begin{equation}
  s(f_i, t) = \sigma\!\left(\cos\bigl(E_V(f_i),\; E_T(t)\bigr)\right),
  \label{eq:siglip_score}
\end{equation}
where $E_V$ and $E_T$ are the vision and text encoders,
$\cos(\cdot,\cdot)$ denotes cosine similarity,
and $\sigma$ is the sigmoid function.

\paragraph{Text modes.}
We employ two modes for $t$: in Q-mode $t$ is the question text, and in
QA-mode $t$ is the concatenation of the question and the correct answer.
QA-mode requires oracle answer labels
and serves as an upper bound on encoder-based selection.

\subsection{Frame Selection Strategies}

Using the frame scores $\{s(f_i, t)\}_{i=1}^L$, we define five different strategies:

\paragraph{\maxprob{}.}
Select the top-$K$ scoring frames:
\begin{equation}
  \mathcal{F}^+ = \operatorname{top\text{-}K}\bigl(f_1, \ldots, f_L;\; s(\cdot, t)\bigr).
\end{equation}

\paragraph{\minprob{}.}
Select the bottom-$K$ scoring frames:
\begin{equation}
  \mathcal{F}^- = \operatorname{bot\text{-}K}\bigl(f_1, \ldots, f_L;\; s(\cdot, t)\bigr).
\end{equation}
This adversarial baseline serves as the diagnostic counterpart to \maxprob{}.

\paragraph{Temporal diversity constraint.}
Adjacent frames within the same shot often receive nearly identical scores,
causing unconstrained top-$K$ selection to concentrate on only a few temporal
regions. This effect is especially severe for \minprob{}, and can therefore be
mistaken for genuine frame sensitivity. To ensure comparable temporal coverage,
we group frames by integer second and select them in rounds. In each round, we
take the next highest-scoring frame from every second, so that the first
$\min(K, \lceil L/\text{fps} \rceil)$ selected frames come from distinct
seconds. \minprob{} uses the same procedure with the ranking reversed.

\paragraph{Uniform.}
Select $K$ frames uniformly distributed over the video timeline.
This is the standard baseline in most deployed systems.

\paragraph{Window oracle.}
Slide a window of $K$ frames with stride $S$ across the video,
run the VLM on each window, and report the best accuracy.
When the total number of windows exceeds a budget $W_{\max}$ (default $50$),
we uniformly subsample $W_{\max}$ windows to bound computation.
This measures the theoretical upper bound achievable
by any temporal selection strategy given window size $K$.

\paragraph{No-frame.}
Replace all input frames with a uniform gray dummy image.
The VLM processes the question but receives no visual content.

\subsection{Frame Selection Sensitivity (FSS)}
\label{ssec:fss}

To measure how much each sample depends on receiving the right frames,
we define \emph{Frame Selection Sensitivity} (\FSS{}).
Let $\mathcal{M}$ denote the set of models, and let
$\text{acc}_{m,i}^+$ and $\text{acc}_{m,i}^-$ be the binary correctness
(1 or 0) of model $m$ on sample $i$
when given \maxprob{} and \minprob{} frames, respectively.
\begin{equation}
  \FSS{}_i = \frac{1}{|\mathcal{M}|}\sum_{m \in \mathcal{M}}
    \bigl(\text{acc}_{m,i}^+ - \text{acc}_{m,i}^-\bigr).
  \label{eq:fss}
\end{equation}
$\FSS{}_i > 0$ indicates that the sample benefits from relevant frames;
$\FSS{}_i < 0$ indicates that relevant frames \emph{hurt}---a visual appearance bias.

\paragraph{Sample categories.}
Given a threshold $\tau$ (default $\tau = 0.15$), we classify each sample as:
\begin{itemize}[topsep=1pt,parsep=0pt,itemsep=1pt]
  \item \emph{Key} ($\FSS{}_i > \tau$): temporally sensitive; benefits from correct frame selection.
  \item \emph{Non-key} ($|\FSS{}_i| \leq \tau$): frame-agnostic; answered regardless of frame choice.
  \item \emph{Anti-key} ($\FSS{}_i < -\tau$): appearance-biased; ``relevant'' frames are detrimental.
\end{itemize}

\noindent
This classification is an \emph{operational} definition of temporal sensitivity:
it captures whether frame selection affects model behavior,
which is a necessary---but not necessarily sufficient---condition
for a question to require temporal reasoning.
\FSS{}-based pruning retains samples with $|\FSS{}| > \tau$,
yielding an evaluation subset
in which frame selection is consequential.

\paragraph{Choice of $\tau$.}
We set $\tau = 0.15$, so that a Key or Anti-key label requires more
than one model's net flip.
Agreement with human temporal judgments and ablations over $\tau$ are in
\Cref{app:human_eval}.

\paragraph{Default configuration.}
We adopt QA-mode as the primary \FSS{} scoring mode,
as it more accurately identifies temporally sensitive questions
by leveraging full question-answer context.
Since \FSS{} is an analytical diagnostic,
maximizing accuracy takes priority over label-free operation.

\subsection{Language Independence Score (LIS)}
\label{ssec:lis}

Let $\text{correct}_{m,i}^{\text{nf}}$ be the binary correctness
of model $m$ on sample $i$ under the no-frame strategy.
\begin{equation}
  \LIS{}_i = 1 - \frac{1}{|\mathcal{M}|}\sum_{m \in \mathcal{M}}
    \text{correct}_{m,i}^{\text{nf}}.
  \label{eq:lis}
\end{equation}
$\LIS{}_i = 1$ when no model answers correctly without meaningful visual input, indicating maximal vision dependence;
$\LIS{}_i = 0$ when all models answer correctly, indicating that the item is solvable from language priors alone.

\section{Experimental Setup}
\label{sec:setup}

\subsection{Models}

We evaluate eight VLMs from five families:
LLaVA-OneVision~\citep{li2024llava},
VideoLLaMA3~\citep{zhang2025videollama},
SmolVLM2~\citep{marafioti2025smolvlm},
Qwen3-VL~\citep{qwen3vl_arxiv2025},
and InternVL3~\citep{zhu2025internvl3},
spanning 256M to 8B parameters.
All models receive $K\!=\!32$ frames in a zero-shot multiple-choice setting;
ablations over $K \in \{16, 32, 48\}$ appear in \Cref{app:ablation}.
We compute frame relevance scores (\Cref{eq:siglip_score})
with SigLIP2-400M~\citep{tschannen2025siglip} as the default; ablations with
SigLIP-400M and CLIP ViT-L/14 appear in \Cref{app:encoder}.
See \Cref{tab:models} for full details.

\subsection{Benchmarks}

We use six Video QA benchmarks grouped by video length.
\emph{Short} ($\leq$3\,min):
MVBench~\citep{li2024mvbench},
NExTQA~\citep{xiao2021next},
and EgoSchema~\citep{mangalam2023egoschema}.
\emph{Long} ($>$5\,min):
MLVU~\citep{mlvu_cvpr2025},
LongVideoBench~\citep{wu2024longvideobench},
and Video-MME~\citep{fu2024videomme}.
Detailed information and full statistics appear in \Cref{tab:benchmarks}.
Design intent and multi-axis linguistic profiles of the six benchmarks
appear in \Cref{app:benchmarks}.

\paragraph{Roadmap.}
We first examine whether frame selection affects VLM accuracy (\S\ref{sec:results}),
then characterize how temporally grounded current benchmarks are (\S\ref{sec:analysis}),
and finally construct \TempCore{}, temporally grounded evaluation subsets (\S\ref{sec:tempcore}).

\section{Does Frame Selection Work?}
\label{sec:results}

\subsection{RQ1: Are Vision Encoders Robust to Interrogative Cues?}
\label{ssec:rq1}

Vision-language similarity is an intuitive basis for frame scoring,
yet VQA queries are inherently interrogative:
they contain words like ``who,'' ``what,'' and ``how many''
that carry little visual grounding.
Since encoders such as SigLIP are trained on declarative image--caption pairs,
their similarity estimates may be poorly calibrated for question-form inputs.
We test this by comparing Q-mode (question only) with QA-mode
(question + ground-truth answer), where the latter supplies
a declarative phrase that better matches the encoder's training distribution.
\Cref{tab:mode_comparison} summarizes benchmark-averaged accuracy.

QA-mode outperforms Q-mode on all six benchmarks,
with an average gain of $+2.3$\% (\Cref{tab:mode_comparison}),
indicating that encoders are weakened by interrogative cues.
The smallest gain among the long-video benchmarks falls to LongVideoBench,
whose questions are up to nearly four times longer \emph{in word count}
than those of other benchmarks
and already contain rich visual descriptions,
making Q-mode nearly sufficient on its own
(see \Cref{app:text_stats} for statistics and examples).

\begin{figure}[t]
  \centering
  \includegraphics[width=\columnwidth]{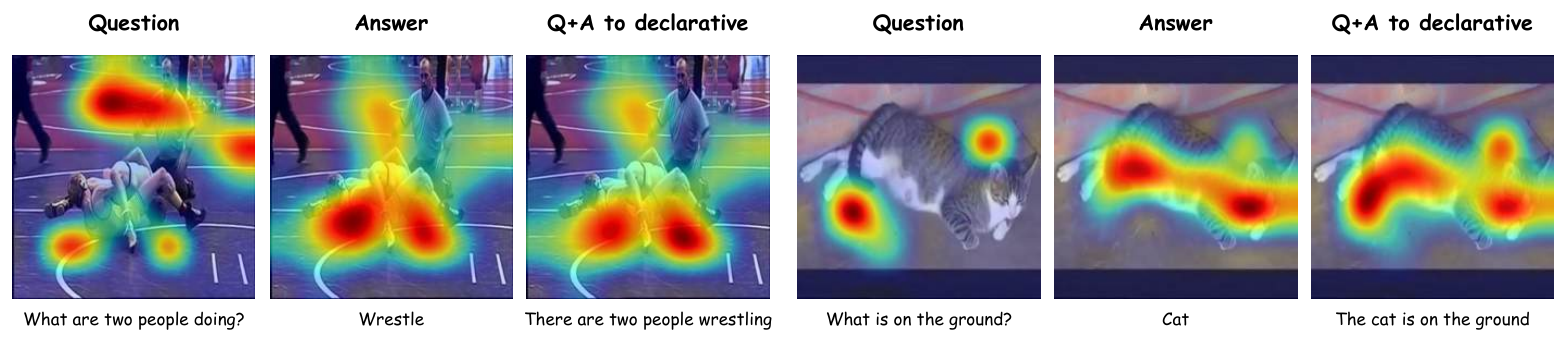}
  \caption{\textbf{Grad-CAM attention maps for question-text vs.\ answer-text scoring.}
    Question-text attention is diffuse,
    while answer-text attention concentrates on the relevant region,
    illustrating the noise introduced by interrogative cues.}
  \label{fig:gradcam}
\end{figure}

\Cref{fig:gradcam} further illustrates this vulnerability via
Grad-CAM~\citep{selvaraju2017_grad} attention maps:
when scoring with question text, attention is diffuse across the frame,
whereas answer text concentrates attention on the task-relevant region.
This suggests that ambiguous, interrogative queries
yield noisier similarity estimates.

\begin{table}[t]
\centering
\small
\caption{\textbf{QA-mode scoring beats Q-mode on all six benchmarks.} \maxprob{} accuracy (\%) averaged across the eight VLMs, with $\Delta$ = QA-mode $-$ Q-mode. Supplying the answer phrase gives the encoder a declarative target closer to its training distribution.}
\label{tab:mode_comparison}
\resizebox{\columnwidth}{!}{%
\begin{tabular}{lrrr}
\toprule
Benchmark & Q-mode & QA-mode & $\Delta$ \\
\midrule
\multicolumn{4}{l}{\textit{Short-video}} \\
MVBench & 48.4 & 49.0 & +0.6 \\
NExTQA & 58.5 & 59.2 & +0.7 \\
EgoSchema & 34.2 & 39.2 & +5.0 \\
\midrule
\multicolumn{4}{l}{\textit{Long-video}} \\
MLVU & 54.9 & 57.3 & +2.4 \\
LongVideoBench & 46.7 & 47.8 & +1.1 \\
Video-MME & 46.0 & 49.7 & +3.7 \\
\midrule
Average & 48.1 & 50.4 & +2.3 \\
\bottomrule
\end{tabular}%
}
\end{table}

\paragraph{Takeaway.}
These results suggest that text-based frame selection methods
must be robust to interrogative and ambiguous input cues.
Developing encoders that handle question-form queries
as effectively as declarative captions
remains an important direction for future work.

\subsection{RQ2: MaxProb vs.\ MinProb}
\label{ssec:rq2}

\begin{figure}[t]
\centering
  \begin{subfigure}{0.49\linewidth}
    \centering
    \includegraphics[width=\linewidth]{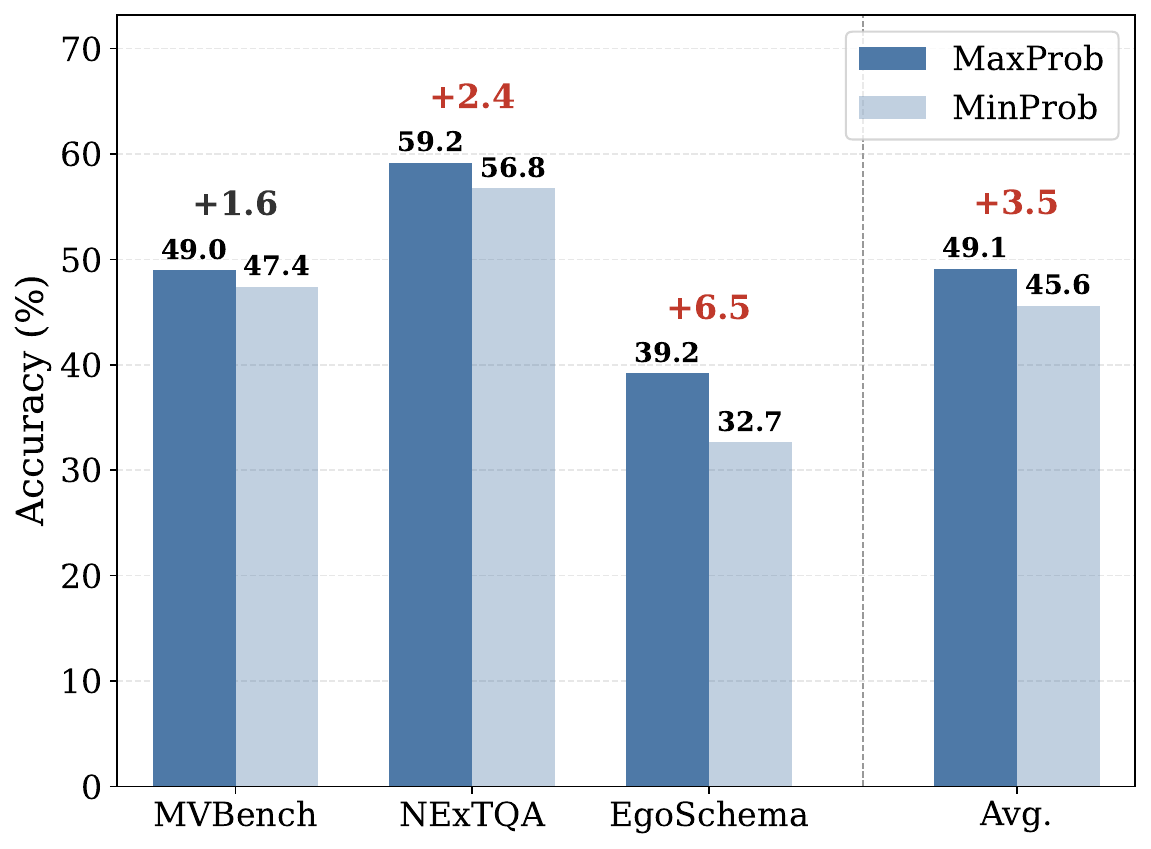}
    \subcaption{Short-video}\label{fig:maxmin-short}
  \end{subfigure}\hfill
  \begin{subfigure}{0.49\linewidth}
    \centering
    \includegraphics[width=\linewidth]{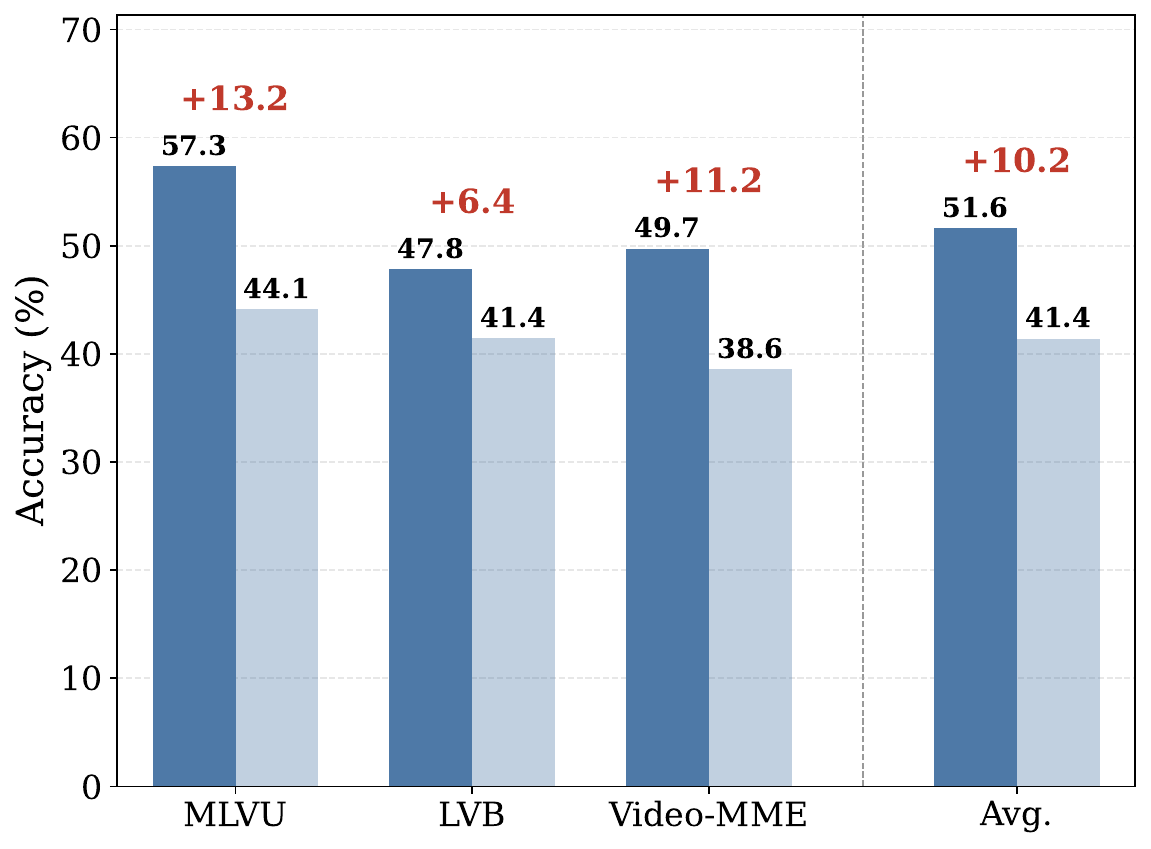}
    \subcaption{Long-video}\label{fig:maxmin-long}
  \end{subfigure}
  \caption{\textbf{Frame selection helps on every benchmark, but three
    times more on long videos.} \maxprob{} vs.\ \minprob{} accuracy
    (8-model average). The mean gap is $+3.5$\,pp on short-video benchmarks
    against $+10.2$\,pp on long-video ones.
    LVB denotes LongVideoBench.}
  \label{fig:maxmin}
\end{figure}

Despite the interrogative-cue limitation identified above,
does vision-language similarity still provide an effective frame selection signal?
\Cref{fig:maxmin} compares \maxprob{} and \minprob{} accuracy
(8-model average) across six benchmarks
(full per-model results in \Cref{tab:maxmin}).

On long-video benchmarks, \maxprob{} consistently outperforms \minprob{},
with an average gain of $+10.2$\% across models.
On short-video benchmarks (MVBench, NExTQA),
the gain is markedly smaller:
in short clips, most $K\!=\!32$ windows already contain the relevant content,
so the scorer's ranking provides little additional benefit.

\paragraph{Takeaway.}
Vision-language similarity is an effective frame selection signal
for long videos, where relevant content is temporally sparse.
On short videos, the benefit shrinks to a few points---temporal information demand
is inherently low, and any $K\!=\!32$ window
already covers most of the relevant content.

\subsection{RQ3: Oracle Gap}
\label{ssec:rq3}

\begin{figure}[t]
\centering
  \begin{subfigure}{0.49\linewidth}
    \centering
    \includegraphics[width=\linewidth]{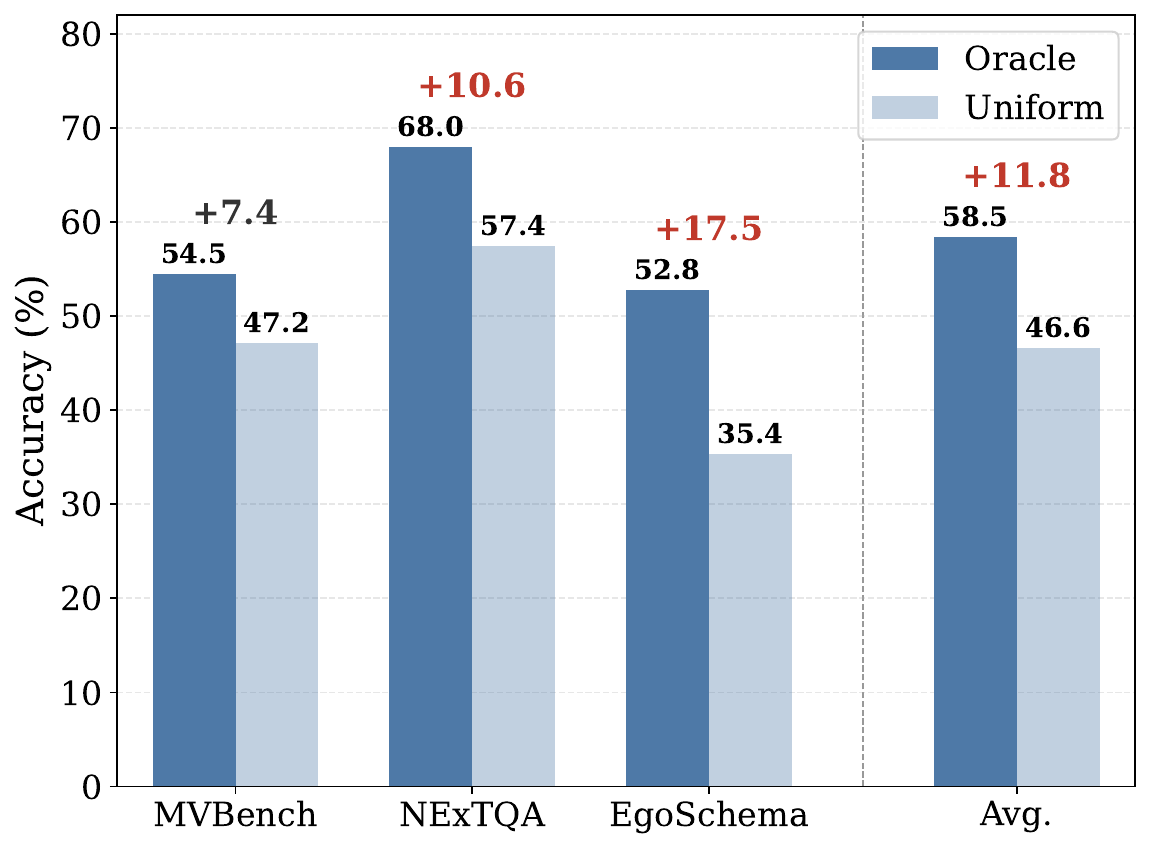}
    \subcaption{Short-video}\label{fig:oracle-short}
  \end{subfigure}\hfill
  \begin{subfigure}{0.49\linewidth}
    \centering
    \includegraphics[width=\linewidth]{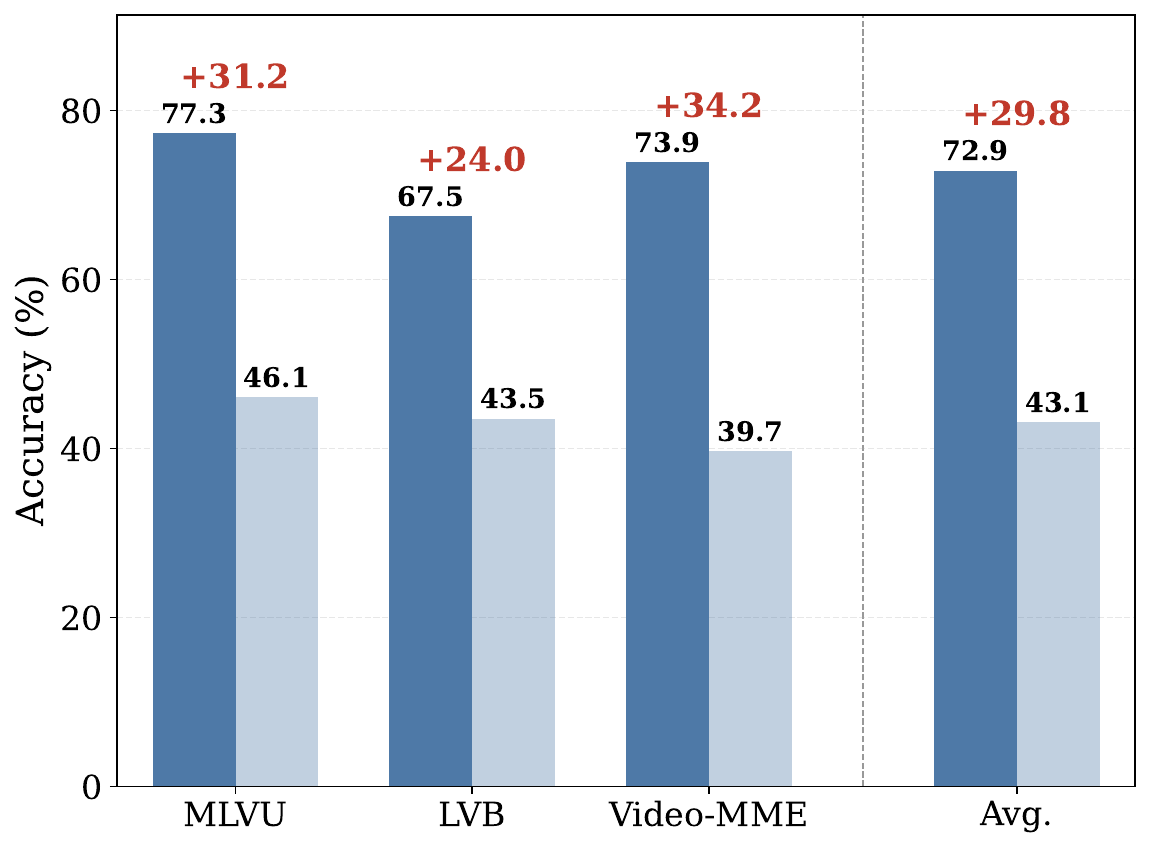}
    \subcaption{Long-video}\label{fig:oracle-long}
  \end{subfigure}
  \caption{\textbf{Oracle gaps reveal large untapped headroom for frame selection.}
    Uniform-sampling vs.\ window-oracle accuracy (model average).
    Long-video gaps far exceed short-video gaps,
    confirming that temporal frame selection is most consequential for extended videos.
    LVB denotes LongVideoBench.}
  \label{fig:oracle}
\end{figure}

How much room remains for better frame selection?
\Cref{fig:oracle} compares window-oracle accuracy---the best accuracy
achievable across all $K$-frame sliding windows---against
a uniform-sampling baseline
(full per-model results in \Cref{tab:oracle}).

The gaps are substantial:
long-video benchmarks average $+25.2$\% while short-video benchmarks
average $+9.9$\%.
However, the oracle selects the single best window per sample,
and different windows may elicit correct answers
through distinct reasoning paths,
potentially inflating the gap beyond pure temporal content.

To control for this concern,
\Cref{tab:oracle_subset} decomposes oracle gaps by \FSS{} category.
Key samples exhibit consistently larger gaps than Non-key samples
on all six benchmarks,
confirming that the more temporal a sample is,
the more headroom remains for improved frame selection.

\begin{table}[t]
\centering
\small
\setlength{\tabcolsep}{4pt}
\resizebox{\columnwidth}{!}{%
\begin{tabular}{llrrrr}
\toprule
Benchmark & Split & $N$ & Uniform & Oracle & Gap \\
\midrule
\multicolumn{6}{l}{\textit{Short-video}} \\
MVBench & All & 3786 & 48.7 & 54.5 & +5.8 \\
 & Key & 212 & 52.2 & 65.4 & \textbf{+13.2} \\
 & Non-key & 3502 & 48.4 & 53.7 & +5.3 \\
 & Anti-key & 72 & 50.9 & 61.3 & +10.4 \\
\midrule
NExTQA & All & 8438 & 58.9 & 68.0 & +9.1 \\
 & Key & 656 & 51.6 & 69.2 & \textbf{+17.5} \\
 & Non-key & 7588 & 59.7 & 67.9 & +8.2 \\
 & Anti-key & 194 & 51.7 & 68.4 & +16.7 \\
\midrule
EgoSchema & All & 500 & 38.1 & 52.8 & +14.7 \\
 & Key & 94 & 41.4 & 61.7 & +20.3 \\
 & Non-key & 390 & 37.1 & 50.2 & +13.1 \\
 & Anti-key & 16 & 43.0 & 64.1 & \textbf{+21.1} \\
\midrule
\multicolumn{6}{l}{\textit{Long-video}} \\
MLVU & All & 2173 & 51.5 & 77.3 & +25.8 \\
 & Key & 782 & 50.6 & 81.1 & \textbf{+30.4} \\
 & Non-key & 1212 & 52.3 & 74.5 & +22.2 \\
 & Anti-key & 179 & 50.1 & 79.8 & +29.7 \\
\midrule
LongVideoBench & All & 1122 & 46.9 & 67.5 & +20.5 \\
 & Key & 229 & 46.7 & 76.1 & \textbf{+29.3} \\
 & Non-key & 823 & 47.2 & 64.9 & +17.7 \\
 & Anti-key & 69 & 44.9 & 71.3 & +26.4 \\
\midrule
Video-MME & All & 2625 & 46.4 & 73.9 & +27.6 \\
 & Key & 783 & 50.9 & 82.6 & +31.6 \\
 & Non-key & 1673 & 44.2 & 69.3 & +25.1 \\
 & Anti-key & 168 & 46.4 & 79.6 & \textbf{+33.2} \\
\bottomrule
\end{tabular}%
}
\caption{\textbf{Key and Anti-key samples exhibit larger oracle gaps than Non-key samples on all six benchmarks.} Oracle gap (\%) decomposed by \FSS{} category, on the samples with both window-oracle and uniform-strategy results. Largest per-benchmark gap in bold, demonstrating that \FSS{}-classified samples are precisely those where temporal frame selection is consequential.}
\label{tab:oracle_subset}
\end{table}

\paragraph{Takeaway.}
Large oracle gaps indicate substantial room for better frame selection,
and this gap is concentrated in temporally sensitive samples.
Longer benchmarks with higher temporal purity
stand to benefit the most from advances in frame retrieval.

\section{How Temporal Are Video QA Benchmarks?}
\label{sec:analysis}

\subsection{FSS Distributions}
\label{ssec:fss_dist}

\begin{figure}[t]
\centering
  \begin{subfigure}{0.49\linewidth}
    \centering
    \includegraphics[width=\linewidth]{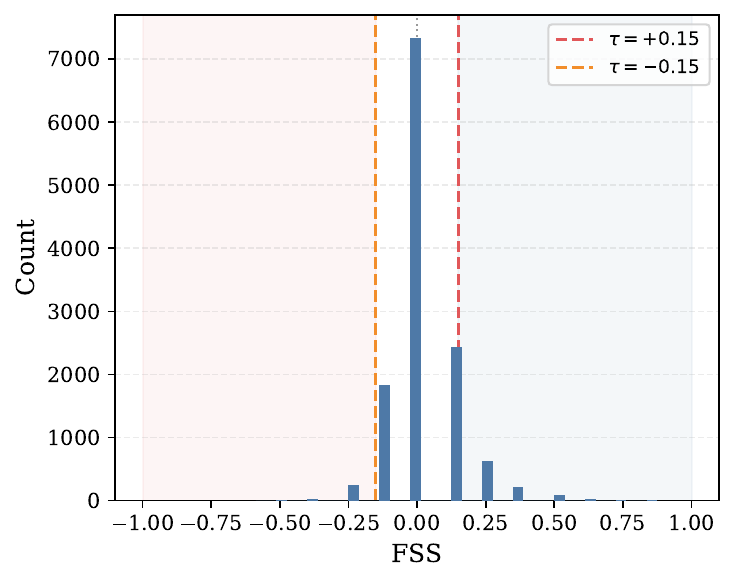}
    \subcaption{Short-video}\label{fig:fss_dist-short}
  \end{subfigure}\hfill
  \begin{subfigure}{0.49\linewidth}
    \centering
    \includegraphics[width=\linewidth]{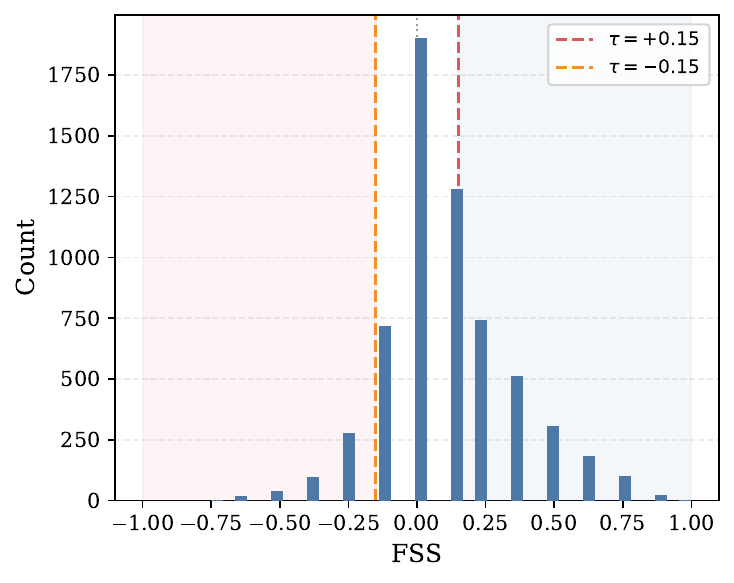}
    \subcaption{Long-video}\label{fig:fss_dist-long}
  \end{subfigure}
  \caption{\textbf{\FSS{} scores concentrate near zero, confirming non-temporal dominance.} Aggregated \FSS{} distributions for short-video and long-video benchmarks. Dashed lines mark classification thresholds ($\tau = \pm 0.15$). The heavy mass near zero shows that frame selection has negligible effect on the majority of samples. Per-benchmark breakdowns are in \Cref{app:fss_detail}.}
  \label{fig:fss_dist}
\end{figure}

\FSS{} is computed from \maxprob{} and \minprob{} accuracy
(defined in \S\ref{sec:method}).
\Cref{fig:fss_dist} shows the distribution of \FSS{} scores across samples
in each benchmark.
The overwhelming majority of samples cluster near zero,
indicating that most Video QA items are insensitive to frame selection.

\paragraph{Quantitative summary.}
\Cref{tab:fss_stats} reports Key and Anti-key fractions for each benchmark.
Key fractions order by video length: MVBench is lowest at $5.6\%$ and MLVU
highest at $36.0\%$, with EgoSchema between the two groups at its intermediate
clip length. Anti-key fractions stay small throughout, so relevant frames help
far more often than they hurt.

\begin{table}[t]
\centering
\setlength{\tabcolsep}{4pt}
\resizebox{0.8\columnwidth}{!}{%
\begin{tabular}{lrrr}
\toprule
Benchmark & Key (\%) & Anti-key (\%) & Key/AK \\
\midrule
\multicolumn{4}{l}{\textit{Short-video}} \\
MVBench      & 5.6 & 1.9 & 2.94 \\
NExTQA       & 7.7 & 2.3 & 3.35 \\
EgoSchema    & 18.8 & 3.2 & 5.88 \\
\midrule
\multicolumn{4}{l}{\textit{Long-video}} \\
MLVU         & \textbf{36.0} & 8.2 & 4.37 \\
LongVideoBench & 21.3 & 6.4 & 3.31 \\
Video-MME    & 29.7 & 6.4 & 4.63 \\
\bottomrule
\end{tabular}%
}
\caption{\textbf{Key fractions order by video length ($\tau=0.15$).} ``Key/AK'' is the ratio of Key to Anti-key samples, above $2.9$ on every benchmark: relevant frames help far more often than they hurt.}
\label{tab:fss_stats}
\end{table}

\subsection{Implications for Benchmark Design}
\label{ssec:bench_design}

Our analysis calls for a reinterpretation of current Video QA benchmarks.
Only a small fraction of samples in any benchmark are temporally sensitive to frame selection;
the remainder are resolved by language priors, visual appearance,
or other non-temporal cues.
This does not invalidate existing benchmarks---appearance and semantic comprehension
are important capabilities---but it underscores that
\emph{temporal} performance should be reported separately.
\FSS{}-based pruning (retaining $|\FSS{}| > \tau$) provides a lightweight, model-agnostic mechanism to do so.
We address this by combining \FSS{} with the Language Independence Score (\LIS{}) to construct \TempCore{}, temporally grounded evaluation subsets.

\section{\TempCore{}}
\label{sec:tempcore}

\begin{figure*}[t]
  \centering
  \includegraphics[width=0.81\textwidth]{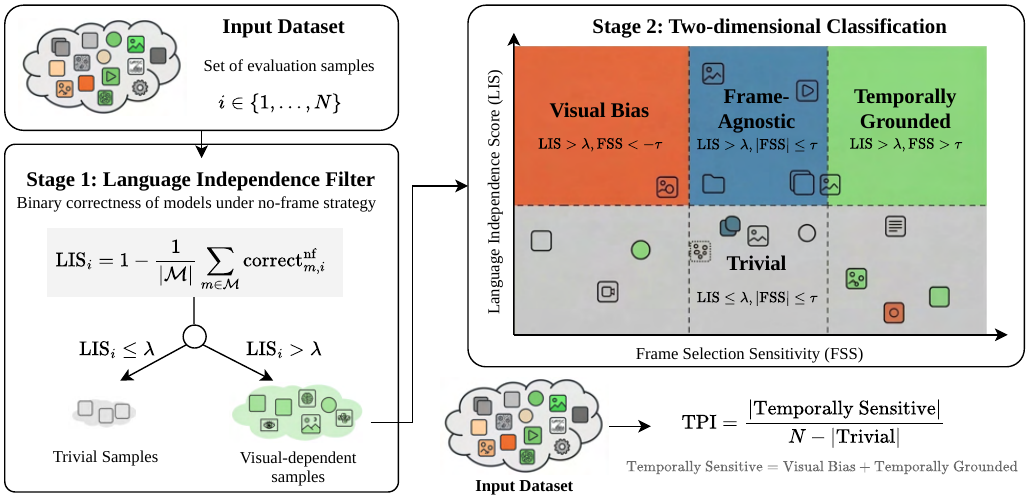}
  \caption{\textbf{Overview of the \TempCore{} construction pipeline.} Stage~1 filters out Trivial samples (solvable from language priors, $\LIS{} \leq \lambda$). Stage~2 classifies the remaining vision-dependent samples along the \FSS{} axis into three categories: Temporally Grounded ($\FSS{} > \tau$), Frame-Agnostic ($|\FSS{}| \leq \tau$), and Visual Bias ($\FSS{} < -\tau$). The Temporal Purity Index (TPI) summarizes the fraction of vision-dependent samples that are temporally sensitive.}
  \label{fig:tempcore_pipeline}
\end{figure*}

\FSS{} captures sensitivity to frame selection
but cannot distinguish genuinely temporal samples
from those solvable by language priors alone---a
high-\FSS{} question may simply contain answer-leaking text cues.
To filter out such cases,
we combine \FSS{} with the Language Independence Score (\LIS{},
introduced in \S\ref{ssec:lis}) and construct
\TempCore{}, temporally grounded evaluation subsets
extracted from existing benchmarks without manual re-annotation
(\Cref{fig:tempcore_pipeline}).

\subsection{Two-Dimensional Taxonomy}
\label{ssec:two_stage}

Combining \LIS{} and \FSS{} yields a three-tier taxonomy, with
Temporally Sensitive samples further divided into two subcategories.
Given thresholds $\tau$ (FSS, default $0.15$) and $\lambda$
(LIS, default $0.50$):
\textbf{Trivial} samples ($\LIS{} \leq \lambda$) are solvable from text alone
and are removed first. Among the remaining vision-dependent samples
($\LIS{} > \lambda$), \textbf{Frame-Agnostic} samples
($|\FSS{}| \leq \tau$) require visual input but are insensitive to which
frames are shown, whereas \textbf{Temporally Sensitive} samples
($|\FSS{}| > \tau$) are both vision-dependent and frame-sensitive.
We further distinguish the latter by the sign of \FSS{}:
\emph{Temporally Grounded} samples ($\FSS{} > \tau$) benefit from relevant
frames, while \emph{Visual Bias} samples ($\FSS{} < -\tau$) are harmed by
them, indicating that visually matching frames can nevertheless be
temporally misleading.

\paragraph{LIS--FSS orthogonality.}
A key design requirement is that \LIS{} and \FSS{} capture distinct properties.
Spearman rank correlations between the two scores are near zero
across all six benchmarks ($|\rho| < 0.10$),
establishing that language independence and frame sensitivity
are empirically orthogonal dimensions
(see \Cref{fig:lis_fss_detail}).

\subsection{Benchmark Quality Profiles}
\label{ssec:bqp}

We summarize each benchmark with the
\textbf{Temporal Purity Index} (TPI):
\begin{equation}
  \text{TPI} = \frac{|\text{Temporally Sensitive}|}{N - |\text{Trivial}|},
  \label{eq:tpi}
\end{equation}
where $\text{Temporally Sensitive} = \text{Key} \cup \text{Anti-key}$
(i.e., $|\FSS{}| > \tau$ and $\LIS{} > \lambda$),
and $\text{Trivial} = \{i : \LIS{}_i \leq \lambda\}$
encompasses all samples solvable from language priors.
TPI thus measures the fraction of \emph{vision-dependent} samples
for which temporal frame selection is consequential.

\begin{table}[t]
\centering
\small
\caption{\textbf{At most a third of any benchmark is both vision-dependent and frame-sensitive.} Benchmark quality profiles under two-stage (LIS, FSS) classification. LIS is computed from no-frame (dummy visual input) predictions. \textit{Trivial}: solvable from text alone (LIS $\leq \lambda$). \textit{Fr.Agn}: vision-dependent but frame-insensitive. \textit{Vis.Bias}: harmed by encoder-selected frames ($\FSS{} < -\tau$). \textit{Temp.Grnd}: benefits from relevant frames ($\FSS{} > \tau$). TempSens = Vis.Bias + Temp.Grnd. TPI = TempSens\,/\,(N $-$ Trivial).}
\label{tab:bqp}
\setlength{\tabcolsep}{3pt}
\resizebox{\columnwidth}{!}{%
\begin{tabular}{lrrrrrr@{\hskip 8pt}r}
\toprule
Benchmark & $N$ & Trivial (\%) & Fr.Agn (\%) & Vis.Bias (\%) & Temp.Grnd (\%) & \textbf{TempSens (\%)} & TPI \\
\midrule
\multicolumn{8}{l}{\textit{Short-video}} \\
MVBench & 3786 & 33.4 & 61.0 & 1.2 & 4.3 & \textbf{5.5} & 0.083 \\
NExTQA & 8564 & 37.0 & 55.7 & 1.5 & 5.8 & \textbf{7.3} & 0.117 \\
EgoSchema & 500 & 20.2 & 60.6 & 2.6 & 16.6 & \textbf{19.2} & 0.241 \\
\midrule
\multicolumn{8}{l}{\textit{Long-video}} \\
MLVU & 2173 & 35.6 & 33.4 & 5.8 & 25.2 & \textbf{31.0} & 0.481 \\
LongVideoBench & 1337 & 36.6 & 43.9 & 4.2 & 15.3 & \textbf{19.5} & 0.308 \\
Video-MME & 2700 & 31.3 & 43.0 & 3.9 & 21.8 & \textbf{25.6} & 0.373 \\
\bottomrule
\end{tabular}%
}
\end{table}

\Cref{tab:bqp} reports the category breakdown.
Several patterns emerge:
(i)~\textbf{Temporally Sensitive} fractions range from $5.5\%$ (MVBench)
to $31.0\%$ (MLVU), indicating that even long-video benchmarks
contain $<\!32\%$ temporally sensitive, vision-dependent samples.
(ii)~\textbf{Frame-Agnostic} samples dominate
MVBench ($61.0\%$) and EgoSchema ($60.6\%$),
suggesting these benchmarks heavily test semantic scene understanding rather than temporal reasoning.
(iii)~\textbf{Trivial} fractions ($20$--$37\%$) are substantial.
(iv)~The Temporal Purity Index ranges from $0.083$ to $0.481$,
providing a single-number summary of each benchmark's temporal grounding quality.

\subsection{Validation}
\label{ssec:tempcore_validation}

Fully validating temporal necessity is inherently difficult, as it is a latent property that cannot be observed directly from annotations alone. We therefore treat validation as converging evidence across complementary signals: human judgments, oracle-gap patterns, and internal and external correlations.

\begin{figure}[t]
  \centering
  \includegraphics[width=0.65\columnwidth]{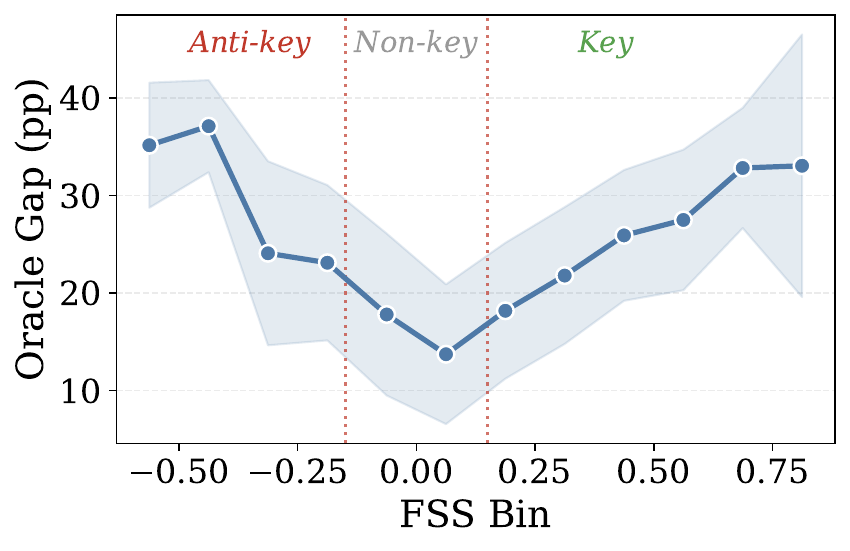}
  \caption{\textbf{Oracle gap exhibits a U-shape across FSS bins (cross-benchmark average).}
  Both Key and Anti-key tails show
  higher oracle gaps than the Non-key center, confirming that
  Anti-key samples are temporally sensitive.
  Shaded region: $\pm 1$ std across benchmarks.}
  \label{fig:oracle_gap_ushape_avg}
\end{figure}

\paragraph{Anti-key temporal dependence.}
A crucial design choice in \TempCore{} is including Anti-key samples
($\FSS{} < -\tau$) alongside Key samples as temporally sensitive.
\Cref{fig:oracle_gap_ushape_avg} aggregates oracle gap across all six benchmarks
by FSS bin, revealing a clear U-shape:
both positive (Key) and negative (Anti-key) tails exhibit elevated oracle gaps
compared to the near-zero center,
confirming that Anti-key samples---like Key samples---benefit
substantially from oracle frame selection.

\begin{figure}[t]
\centering
  \begin{subfigure}{0.49\linewidth}
    \centering
    \includegraphics[width=\linewidth]{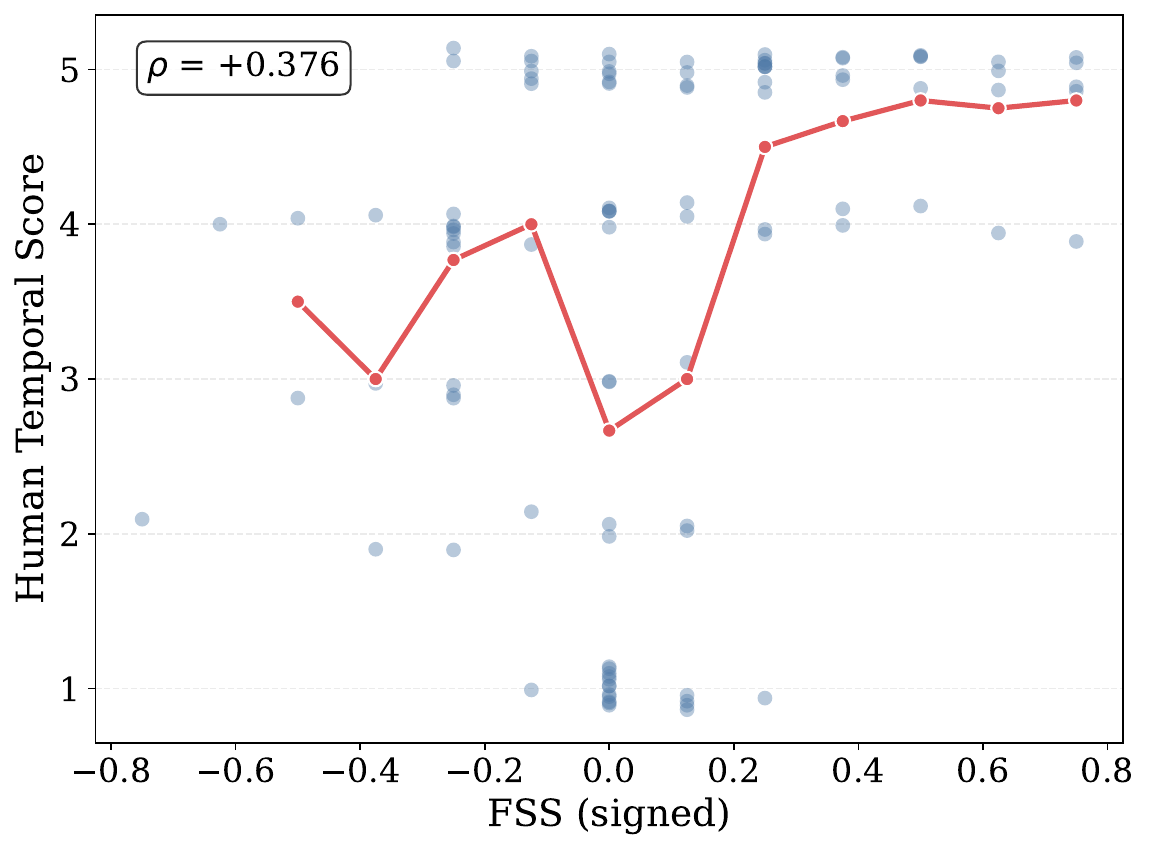}
    \subcaption{FSS ($\rho = 0.376$)}\label{fig:human_fss_scatter-signed}
  \end{subfigure}\hfill
  \begin{subfigure}{0.49\linewidth}
    \centering
    \includegraphics[width=\linewidth]{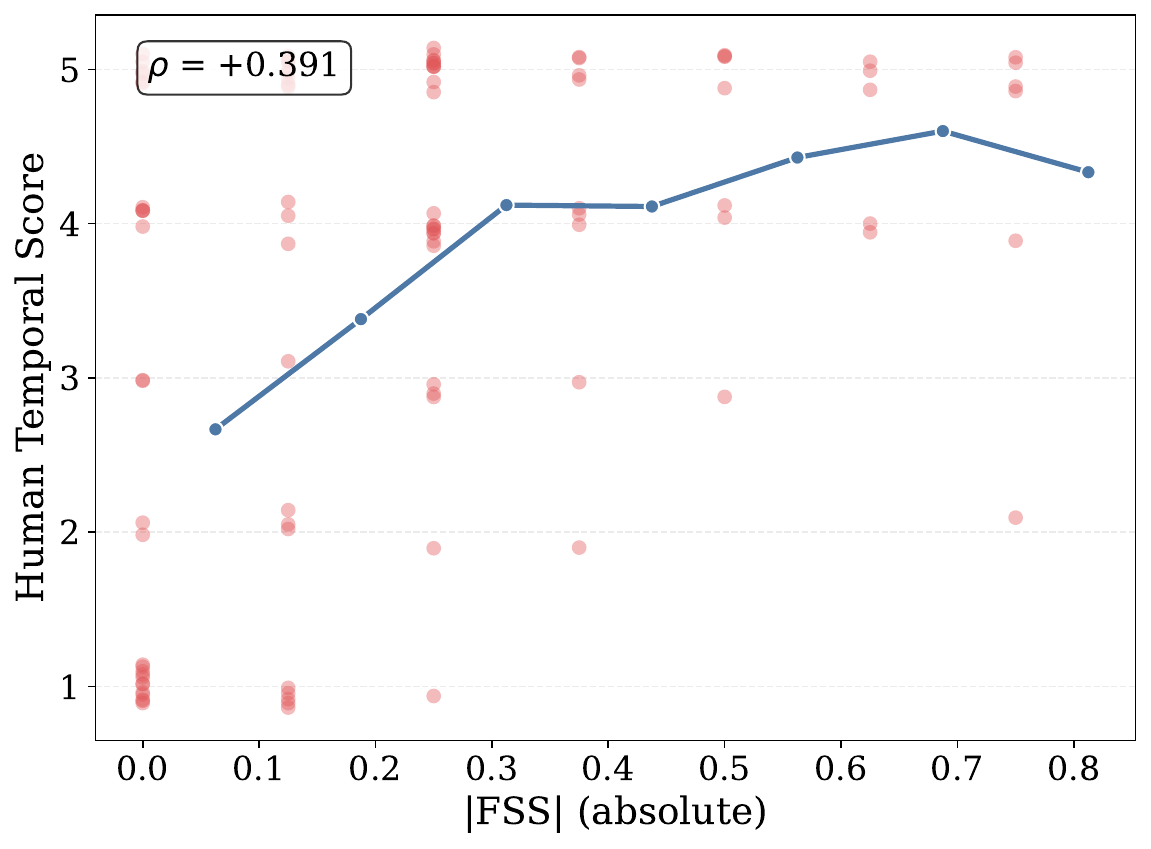}
    \subcaption{$|$FSS$|$ ($\rho = 0.391$)}\label{fig:human_fss_scatter-abs}
  \end{subfigure}
  \caption{\textbf{FSS vs.\ human score.}
  (a)~FSS shows positive correlation.
  (b)~$|$FSS$|$ yields stronger correlation,
  capturing temporal dependence at both extremes.}
  \label{fig:human_fss_scatter}
\end{figure}

Human evaluation corroborates this finding.
\Cref{fig:human_fss_scatter} shows that $|\text{FSS}|$ achieves a stronger correlation with human temporal scores
than FSS ($\rho = +0.391$ vs.\ $+0.376$),
and Anti-key samples receive higher temporal ratings
than Non-key samples ($3.55$ vs.\ $2.98$).
Anti-key samples are visually misled but temporally dependent:
SigLIP's highest-confidence frames capture appearance rather than temporal content,
and removing them paradoxically helps
because the model falls back on more temporally informative frames.
See additional analysis and details in \Cref{app:validation}.

\begin{figure}[t]
  \centering
  \includegraphics[width=0.65\columnwidth]{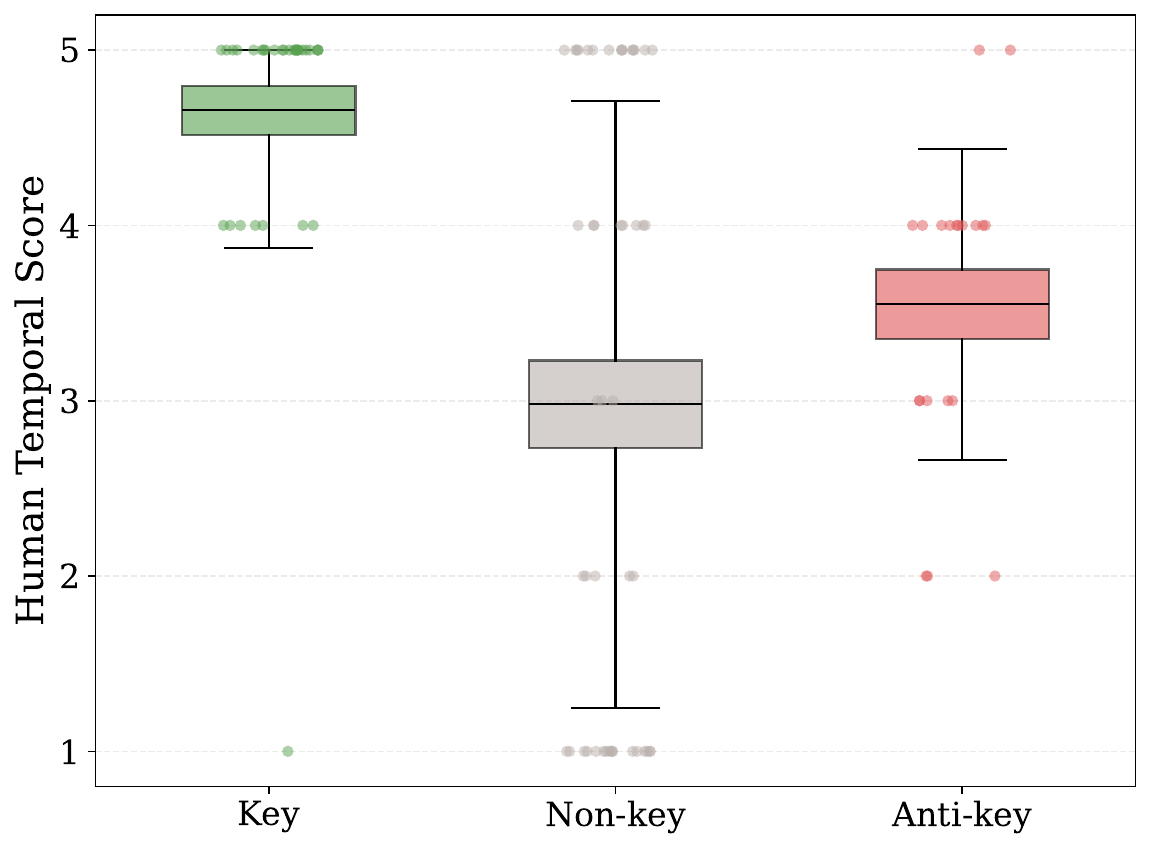}
  \caption{\textbf{Human temporal scores by FSS category ($\tau = 0.15$).}
  Key samples receive the highest ratings;
  Anti-key exceeds Non-key, supporting their inclusion
  in Temporally Sensitive.
  Boxes: mean $\pm$ 1\,SE; whiskers: $\pm$ 1\,SD; centre line marks the mean.}
  \label{fig:human_fss_bin_boxplot}
\end{figure}

\paragraph{Internal consistency.}
Oracle gaps on Key samples exceed those on Non-key samples
on all six benchmarks
(mean $+23.7$ vs.\ $+15.3$; \Cref{tab:oracle_subset}),
and human annotators rate Key samples highest, followed by Anti-key and Non-key (see \Cref{fig:human_fss_bin_boxplot}),
confirming that \TempCore{} retains those samples
where temporal frame selection is consequential.

\paragraph{External correlation.}
Non-key model rankings correlate more strongly with
six image VLM benchmarks than Key rankings
(mean $\rho = 0.79$ vs.\ $0.59$ across benchmarks;
\Cref{app:external_corr}),
confirming that Key performance captures a distinct temporal skill
rather than general VLM capability.

\paragraph{Practical usage.}
\TempCore{} subsets contain samples with $|\FSS{}| > \tau$ and $\LIS{} > \lambda$,
comprising $5.5$--$31.0$\% of the original evaluation set.
They can be used as a complementary metric:
reporting \TempCore{}-subset accuracy alongside overall benchmark scores
separates frame-sensitive from frame-agnostic performance.
The pruning removes the $69$--$94$\% of samples that are
not frame-sensitive,
yielding a focused evaluation of frame-selection sensitivity.

\section{Conclusion}
\label{sec:conclusion}

We presented three contributions toward understanding
temporal grounding in Video QA evaluation.
First, we proposed \FSS{}, a per-sample diagnostic that quantifies
dependence on temporal frame selection, and conducted a large-scale
systematic study across multiple models, benchmarks, and strategies.
Encoder-based frame selection yields substantial gains on long-video benchmarks,
while oracle ceilings confirm large untapped headroom.
Second, combining \FSS{} with a \LIS{} reveals that only a minority of benchmark samples are temporally sensitive---both
vision-dependent and frame-sensitive.
Third, we constructed \TempCore{}, temporally grounded evaluation subsets, providing a focused measure
of frame-selection sensitivity.

\section*{Limitations}
\label{sec:limitations}

While \FSS{} effectively identifies frame-sensitive samples,
it captures only one facet of temporal reasoning---dependence
on which frames are selected.
Other temporal requirements, such as causal ordering,
event duration estimation, or state-change detection,
may not manifest as frame-selection sensitivity
and are therefore invisible to \FSS{}.
Incorporating complementary temporal diagnostics
(\eg{}, temporal ordering probes, action phase segmentation metrics)
could yield a more comprehensive characterization
of benchmark temporality.
We leave the development of such multi-signal temporal taxonomies
to future work.

We therefore do not claim that \TempCore{} measures temporal reasoning.
A \TempCore{} subset isolates the samples whose answers depend on which frames
are shown---a necessary condition for temporal reasoning (\S\ref{sec:method}),
but not a sufficient one.

\section*{Acknowledgments}

This work was supported in part by the Institute of Information \& Communications Technology Planning \& Evaluation (IITP) grants funded by the Korea government (MSIT) in 2026
(Grant No. RS-2026-25531289, Project Title: Development of Incrementally Adaptive On-device Learning Technology for New Environments, Contribution: 40\%;
Grant No. RS-2024-00457882, Project Title: AI Research Hub Project, Contribution: 10\%;
Grant No. RS-2019-II191906, Project Title: Artificial Intelligence Graduate School Program (POSTECH), Contribution: 10\%)
and in part by the National Research Foundation of Korea (NRF) grants funded by the Korea government (MSIT) in 2026
(Grant No. RS-2026-25494004, Project Title: Extremely Efficient Model Optimization for Large Multimodal Reasoning Foundation Models, Contribution: 20\%;
Grant No. RS-2024-00453301, Project Title: MASC: Convergence of Multimodal AI and Semantic Communications for Hyper-Connected Hyper-Intelligent Public Safety Systems, Contribution: 20\%).

\bibliography{custom}

@STRING{cvpr    = "IEEE/CVF Conference on Computer Vision and Pattern Recognition (CVPR)"}

@STRING{iccv    = "IEEE/CVF International Conference on Computer Vision (ICCV)"}

@STRING{neurips = "Advances in Neural Information Processing Systems (NeurIPS)"}

@STRING{icml    = "International Conference on Machine Learning (ICML)"}

@STRING{iclr    = "International Conference on Learning Representations (ICLR)"}

@STRING{emnlp   = "Empirical Methods in Natural Language Processing (EMNLP)"}

@STRING{acl     = "Proceedings of the Annual Meeting of the Association for Computational Linguistics (ACL)"}

@STRING{aclf    = "Findings of the Association for Computational Linguistics"}

@STRING{bmvc    = "British Machine Vision Conference (BMVC)"}

@inproceedings{frame_voyager_iclr2025,
  title={{Frame-Voyager}: Learning to Query Frames for Video Large Language Models},
  author={Yu, Sicheng and Jin, Chengkai and Wang, Huanyu and Chen, Zhenghao and Jin, Sheng and Zuo, Zhongrong and Xu, Xiaolei and Sun, Zhenbang and Zhang, Bingni and Wu, Jiawei and Zhang, Hao and Sun, Qianru},
  booktitle=iclr,
  year={2025},
}

@inproceedings{mllm_frame_selection_cvpr2025,
  title={{M-LLM} Based Video Frame Selection for Efficient Video Understanding},
  author={Hu, Kai and Gao, Feng and Nie, Xiaohan and Zhou, Peng and Tran, Son and Neiman, Tal and Wang, Lingyun and Shah, Mubarak and Hamid, Raffay and Yin, Bing and Chilimbi, Trishul},
  booktitle=cvpr,
  year={2025},
}

@inproceedings{ffs_cvpr2025,
  title={Flexible Frame Selection for Efficient Video Reasoning},
  author={Buch, Shyamal and Nagrani, Arsha and Arnab, Anurag and Schmid, Cordelia},
  booktitle=cvpr,
  year={2025},
}

@inproceedings{apollo_cvpr2025,
  title={{Apollo}: An Exploration of Video Understanding in Large Multimodal Models},
  author={Zohar, Orr and Wang, Xiaohan and Dubois, Yann and Mehta, Nikhil and Xiao, Tong and Hansen-Estruch, Philippe and Yu, Licheng and Wang, Xiaofang and Juefei-Xu, Felix and Zhang, Ning and Yeung-Levy, Serena and Xia, Xide},
  booktitle=cvpr,
  year={2025},
}

@inproceedings{consistency_videollm_cvpr2025,
  title={On the Consistency of Video Large Language Models in Temporal Comprehension},
  author={Jung, Minjoon and Xiao, Junbin and Zhang, Byoung-Tak and Yao, Angela},
  booktitle=cvpr,
  year={2025},
}

@inproceedings{tvbench_bmvc2025,
  title={Lost in Time: A New Temporal Benchmark for {VideoLLMs}},
  author={Cores, Daniel and Dorkenwald, Michael and Mucientes, Manuel and Snoek, Cees G. M. and Asano, Yuki M.},
  booktitle=bmvc,
  year={2025},
}

@article{frame_sampling_strategies_arxiv2025,
  title={Frame Sampling Strategies Matter: A Benchmark for Small Vision Language Models},
  author={Brkic, Marija and Razzouki, Anas Filali and Tevissen, Yannis and Guetari, Khalil and El Yacoubi, Mounim A.},
  journal={arXiv preprint arXiv:2509.14769},
  year={2025},
}

@inproceedings{mdp3_iccv2025,
  title={{MDP3}: A Training-Free Approach for List-wise Frame Selection in Video-{LLMs}},
  author={Sun, Hui and Lu, Shiyin and Wang, Huanyu and Chen, Qing-Guo and Xu, Zhao and Luo, Weihua and Zhang, Kaifu and Li, Ming},
  booktitle=iccv,
  year={2025},
}

@inproceedings{bootstrap_ci_arxiv2025,
  title={When +1\% is Not Enough: A Paired Bootstrap Protocol for Evaluating Small Improvements},
  author={Du, Wenzhang},
  booktitle={Pacific-Asia Conference on Knowledge Discovery and Data Mining},
  pages={519--531},
  year={2026},
  organization={Springer}
}

@article{text_bias_mcqa_arxiv2026,
  title={Reducing Text Bias in Synthetically Generated {MCQAs} for {VLMs} in Autonomous Driving},
  author={Kulgod, Sutej and Ye, Sean and Tanwar, Sanchit and Heckman, Christoffer},
  journal={arXiv preprint arXiv:2602.17677},
  year={2026},
}

@inproceedings{time_blindness_arxiv2025,
  title={Time Blindness: Why Video-Language Models Can't See What Humans Can?},
  author={Upadhyay, Ujjwal and Ranjan, Mukul and Shen, Zhiqiang and Elhoseiny, Mohamed},
  booktitle=cvpr,
  year={2026}
}

@inproceedings{videotree_cvpr2025,
  title={{VideoTree}: Adaptive Tree-based Video Representation for {LLM} Reasoning on Long Videos},
  author={Wang, Ziyang and Yu, Shoubin and Stengel-Eskin, Elias and Yoon, Jaehong and Cheng, Feng and Bertasius, Gedas and Bansal, Mohit},
  booktitle=cvpr,
  year={2025},
}

@inproceedings{gens_acl2025,
  title={Generative Frame Sampler for Long Video Understanding},
  author={Yao, Linli and Wu, Haoning and Ouyang, Kun and Zhang, Yuanxing and Xiong, Caiming and Chen, Bei and Sun, Xu and Li, Junnan},
  booktitle=aclf,
  year={2025},
}

@article{zhang2025videollama,
  title={{VideoLLaMA3}: Frontier Multimodal Foundation Models for Image and Video Understanding},
  author={Zhang, Boqiang and Li, Kehan and Cheng, Zesen and Hu, Zhiqiang and Yuan, Yuqian and Chen, Guanzheng and Leng, Sicong and Jiang, Yuming and Zhang, Hang and Li, Xin and Jin, Peng and Zhang, Wenqi and Wang, Fan and Bing, Lidong and Zhao, Deli},
  journal={arXiv preprint arXiv:2501.13106},
  year={2025},
}

@article{qwen25vl_arxiv2025,
  title={{Qwen2.5-VL} Technical Report},
  author={Bai, Shuai and Chen, Keqin and Liu, Xuejing and Wang, Jialin and Ge, Wenbin and Song, Sibo and Dang, Kai and Wang, Peng and Wang, Shijie and Tang, Jun and Zhong, Humen and Zhu, Yuanzhi and Yang, Mingkun and Li, Zhaohai and Wan, Jianqiang and others},
  journal={arXiv preprint arXiv:2502.13923},
  year={2025},
}

@article{qwen3vl_arxiv2025,
  title={{Qwen3-VL} Technical Report},
  author={Bai, Shuai and Cai, Yuxuan and Chen, Ruizhe and Chen, Keqin and Chen, Xionghui and Cheng, Zesen and Deng, Lianghao and Ding, Wei and Gao, Chang and Ge, Chunjiang and Ge, Wenbin and Guo, Zhifang and Huang, Qidong and Huang, Jie and Huang, Fei and others},
  journal={arXiv preprint arXiv:2511.21631},
  year={2025},
}

@inproceedings{counterfactual_vqa_cvpr2021,
  title={Counterfactual {VQA}: A Cause-Effect Look at Language Bias},
  author={Niu, Yulei and Tang, Kaihua and Zhang, Hanwang and Lu, Zhiwu and Hua, Xian-Sheng and Wen, Ji-Rong},
  booktitle=cvpr,
  year={2021},
}

@inproceedings{mangalam2023egoschema,
  title={{EgoSchema}: A Diagnostic Benchmark for Very Long-form Video Language Understanding},
  author={Mangalam, Karttikeya and Akshulakov, Raiymbek and Malik, Jitendra},
  booktitle=neurips,
  year={2023},
}

@inproceedings{li2024mvbench,
  title={{MVBench}: A Comprehensive Multi-modal Video Understanding Benchmark},
  author={Li, Kunchang and Wang, Yali and He, Yinan and Li, Yizhuo and Wang, Yi and Liu, Yi and Wang, Zun and Xu, Jilan and Chen, Guo and Luo, Ping and Wang, Limin and Qiao, Yu},
  booktitle=cvpr,
  year={2024},
}

@inproceedings{xiao2021next,
  title={{NExT-QA}: Next Phase of Question-Answering to Explaining Temporal Actions},
  author={Xiao, Junbin and Shang, Xindi and Yao, Angela and Chua, Tat-Seng},
  booktitle=cvpr,
  year={2021},
}

@inproceedings{mlvu_cvpr2025,
  title={Mlvu: Benchmarking multi-task long video understanding},
  author={Zhou, Junjie and Shu, Yan and Zhao, Bo and Wu, Boya and Liang, Zhengyang and Xiao, Shitao and Qin, Minghao and Yang, Xi and Xiong, Yongping and Zhang, Bo and others},
  booktitle=cvpr,
  year={2025},
  organization={IEEE}
}

@inproceedings{wu2024longvideobench,
  title={{LongVideoBench}: A Benchmark for Long-context Interleaved Video-Language Understanding},
  author={Wu, Haoning and Li, Dongxu and Chen, Bei and Li, Junnan},
  booktitle=neurips,
  year={2024},
}

@inproceedings{fu2024videomme,
  title={Video-mme: The first-ever comprehensive evaluation benchmark of multi-modal llms in video analysis},
  author={Fu, Chaoyou and Dai, Yuhan and Luo, Yongdong and Li, Lei and Ren, Shuhuai and Zhang, Renrui and Wang, Zihan and Zhou, Chenyu and Shen, Yunhang and Zhang, Mengdan and others},
  booktitle=cvpr,
  year={2025},
  organization={IEEE}
}

@article{
li2024llava,
title={{LL}a{VA}-OneVision: Easy Visual Task Transfer},
author={Bo Li and Yuanhan Zhang and Dong Guo and Renrui Zhang and Feng Li and Hao Zhang and Kaichen Zhang and Peiyuan Zhang and Yanwei Li and Ziwei Liu and Chunyuan Li},
journal={Transactions on Machine Learning Research},
issn={2835-8856},
year={2025},
url={https://openreview.net/forum?id=zKv8qULV6n},
note={}
}

@inproceedings{
marafioti2025smolvlm,
title={Smol{VLM}: Redefining small and efficient multimodal models},
author={Andr{\'e}s Marafioti and Orr Zohar and Miquel Farr{\'e} and Merve noyan and Elie Bakouch and Pedro Manuel Cuenca Jim{\'e}nez and Cyril Zakka and Loubna Ben allal and Anton Lozhkov and Nouamane Tazi and Vaibhav Srivastav and Joshua Lochner and Hugo Larcher and Mathieu Morlon and Lewis Tunstall and Leandro Von Werra and Thomas Wolf},
booktitle={Second Conference on Language Modeling},
year={2025},
url={https://openreview.net/forum?id=qMUbhGUFUb}
}

@article{zhu2025internvl3,
  title={{InternVL3}: Exploring Advanced Training and Test-Time Recipes for Open-Source Multimodal Models},
  author={Zhu, Jinguo and Wang, Weiyun and Chen, Zhe and Liu, Zhaoyang and Ye, Shenglong and Gu, Lixin and Tian, Hao and Duan, Yuchen and Su, Weijie and Shao, Jie and Gao, Zhangwei and Cui, Erfei and Wang, Xuehui and Cao, Yue and Liu, Yangzhou and others},
  journal={arXiv preprint arXiv:2504.10479},
  year={2025},
}

@inproceedings{zhai2023sigmoid,
  title={Sigmoid Loss for Language Image Pre-Training},
  author={Zhai, Xiaohua and Mustafa, Basil and Kolesnikov, Alexander and Beyer, Lucas},
  booktitle=iccv,
  year={2023},
}

@article{tschannen2025siglip,
  title={{SigLIP 2}: Multilingual Vision-Language Encoders with Improved Semantic Understanding, Localization, and Dense Features},
  author={Tschannen, Michael and Gritsenko, Alexey and Wang, Xiao and Naeem, Muhammad Ferjad and Alabdulmohsin, Ibrahim and Parthasarathy, Nikhil and Evans, Talfan and Beyer, Lucas and Xia, Ye and Mustafa, Basil and H{\'e}naff, Olivier and Harmsen, Jeremiah and Steiner, Andreas and Zhai, Xiaohua},
  journal={arXiv preprint arXiv:2502.14786},
  year={2025},
}

@inproceedings{clip_icml2021,
  title={Learning Transferable Visual Models From Natural Language Supervision},
  author={Radford, Alec and Kim, Jong Wook and Hallacy, Chris and Ramesh, Aditya and Goh, Gabriel and Agarwal, Sandhini and Sastry, Girish and Askell, Amanda and Mishkin, Pamela and Clark, Jack and Krueger, Gretchen and Sutskever, Ilya},
  booktitle=icml,
  year={2021},
}

@inproceedings{lei2023revealing,
  title={Revealing Single Frame Bias for Video-and-Language Learning},
  author={Lei, Jie and Berg, Tamara and Bansal, Mohit},
  booktitle=acl,
  year={2023},
}

@inproceedings{tang2025adaptive,
  title={Adaptive Keyframe Sampling for Long Video Understanding},
  author={Tang, Xi and Qiu, Jihao and Xie, Lingxi and Tian, Yunjie and Jiao, Jianbin and Ye, Qixiang},
  booktitle=cvpr,
  year={2025},
}

@inproceedings{himakunthala2023let,
  title={Let's Think Frame by Frame with {VIP}: A Video Infilling and Prediction Dataset for Evaluating Video Chain-of-Thought},
  author={Himakunthala, Vaishnavi and Ouyang, Andy and Rose, Daniel and He, Ryan and Mei, Alex and Lu, Yujie and Sonar, Chinmay and Saxon, Michael and Wang, William Yang},
  booktitle=emnlp,
  year={2023},
}

@inproceedings{shen2024longvu,
  title={{LongVU}: Spatiotemporal Adaptive Compression for Long Video-Language Understanding},
  author={Shen, Xiaoqian and Xiong, Yunyang and Zhao, Changsheng and Wu, Lemeng and Chen, Jun and Zhu, Chenchen and Liu, Zechun and Xiao, Fanyi and Varadarajan, Balakrishnan and Bordes, Florian and Liu, Zhuang and Xu, Hu and Kim, Hyunwoo J. and Soran, Bilge and Krishnamoorthi, Raghuraman and others},
  booktitle=icml,
  year={2025},
}

@inproceedings{selvaraju2017_grad,
  title={{Grad-CAM}: Visual Explanations from Deep Networks via Gradient-Based Localization},
  author={Selvaraju, Ramprasaath R. and Cogswell, Michael and Das, Abhishek and Vedantam, Ramakrishna and Parikh, Devi and Batra, Dhruv},
  booktitle=iccv,
  year={2017},
}

@inproceedings{
temporalbench_neurips2024,
title={TemporalBench: Benchmarking Fine-grained Temporal Understanding for Multimodal Video Models},
author={Mu Cai and Reuben Tan and Jianrui Zhang and Bocheng Zou and Kai Zhang and Yao Feng and Fangrui Zhu and Jing Gu and Yiwu Zhong and Yuzhang Shang and Yao Dou and Jaden Park and Jianfeng Gao and Yong Jae Lee and Jianwei Yang},
booktitle={Workshop on Video-Language Models @ NeurIPS 2024},
url={https://openreview.net/forum?id=bhuZn4URBH}
}

@inproceedings{liu2024tempcompass,
  title={{TempCompass}: Do Video {LLMs} Really Understand Videos?},
  author={Liu, Yuanxin and Li, Shicheng and Liu, Yi and Wang, Yuxiang and Ren, Shuhuai and Li, Lei and Chen, Sishuo and Sun, Xu and Hou, Lu},
  booktitle=aclf,
  year={2024},
}

@inproceedings{lmmseval2024,
  title={Lmms-eval: Reality check on the evaluation of large multimodal models},
  author={Zhang, Kaichen and Li, Bo and Zhang, Peiyuan and Pu, Fanyi and Cahyono, Joshua Adrian and Hu, Kairui and Liu, Shuai and Zhang, Yuanhan and Yang, Jingkang and Li, Chunyuan and others},
  booktitle={Findings of the Association for Computational Linguistics: NAACL 2025},
  year={2025}
}

@article{achiam2023gpt,
  title={Gpt-4 technical report},
  author={Achiam, Josh and Adler, Steven and Agarwal, Sandhini and Ahmad, Lama and Akkaya, Ilge and Aleman, Florencia Leoni and Almeida, Diogo and Altenschmidt, Janko and Altman, Sam and Anadkat, Shyamal and others},
  journal={arXiv preprint arXiv:2303.08774},
  year={2023}
}

@article{grattafiori2024llama,
  title={The {Llama 3} Herd of Models},
  author={Grattafiori, Aaron and Dubey, Abhimanyu and Jauhri, Abhinav and Pandey, Abhinav and Kadian, Abhishek and Al-Dahle, Ahmad and Letman, Aiesha and Mathur, Akhil and Schelten, Alan and Vaughan, Alex and Yang, Amy and Fan, Angela and Goyal, Anirudh and Hartshorn, Anthony and Yang, Aobo and others},
  journal={arXiv preprint arXiv:2407.21783},
  year={2024},
}

@article{yang2024qwen2,
  title={{Qwen2} Technical Report},
  author={Yang, An and Yang, Baosong and Hui, Binyuan and Zheng, Bo and Yu, Bowen and Zhou, Chang and Li, Chengpeng and Li, Chengyuan and Liu, Dayiheng and Huang, Fei and Dong, Guanting and Wei, Haoran and Lin, Huan and Tang, Jialong and Wang, Jialin and others},
  journal={arXiv preprint arXiv:2407.10671},
  year={2024},
}

@article{team2025gemma,
  title={{Gemma 2}: Improving Open Language Models at a Practical Size},
  author={{Gemma Team} and Riviere, Morgane and Pathak, Shreya and Sessa, Pier Giuseppe and Hardin, Cassidy and Bhupatiraju, Surya and Hussenot, L{\'e}onard and Mesnard, Thomas and Shahriari, Bobak and Ram{\'e}, Alexandre and Ferret, Johan and Liu, Peter and Tafti, Pouya and Friesen, Abe and Casbon, Michelle and others},
  journal={arXiv preprint arXiv:2408.00118},
  year={2024},
}

@inproceedings{alayrac2022flamingo,
  title={{Flamingo}: A Visual Language Model for Few-Shot Learning},
  author={Alayrac, Jean-Baptiste and Donahue, Jeff and Luc, Pauline and Miech, Antoine and Barr, Iain and Hasson, Yana and Lenc, Karel and Mensch, Arthur and Millican, Katherine and Reynolds, Malcolm and Ring, Roman and Rutherford, Eliza and Cabi, Serkan and Han, Tengda and Gong, Zhitao and others},
  booktitle=neurips,
  year={2022},
}

@inproceedings{liu2023visual,
  title={Visual Instruction Tuning},
  author={Liu, Haotian and Li, Chunyuan and Wu, Qingyang and Lee, Yong Jae},
  booktitle=neurips,
  year={2023},
}

@article{chu2024qwen2,
  title={{Qwen2-Audio} Technical Report},
  author={Chu, Yunfei and Xu, Jin and Yang, Qian and Wei, Haojie and Wei, Xipin and Guo, Zhifang and Leng, Yichong and Lv, Yuanjun and He, Jinzheng and Lin, Junyang and Zhou, Chang and Zhou, Jingren},
  journal={arXiv preprint arXiv:2407.10759},
  year={2024},
}

@article{bai2025qwen2,
  title={{Qwen2-VL}: Enhancing Vision-Language Model's Perception of the World at Any Resolution},
  author={Wang, Peng and Bai, Shuai and Tan, Sinan and Wang, Shijie and Fan, Zhihao and Bai, Jinze and Chen, Keqin and Liu, Xuejing and Wang, Jialin and Ge, Wenbin and Fan, Yang and Dang, Kai and Du, Mengfei and Ren, Xuancheng and Men, Rui and others},
  journal={arXiv preprint arXiv:2409.12191},
  year={2024},
}

@inproceedings{zhang2023video,
  title={{Video-LLaMA}: An Instruction-tuned Audio-Visual Language Model for Video Understanding},
  author={Zhang, Hang and Li, Xin and Bing, Lidong},
  booktitle={Proceedings of the 2023 Conference on Empirical Methods in Natural Language Processing: System Demonstrations},
  year={2023},
}

@inproceedings{lin2024video,
  title={{Video-LLaVA}: Learning United Visual Representation by Alignment Before Projection},
  author={Lin, Bin and Ye, Yang and Zhu, Bin and Cui, Jiaxi and Ning, Munan and Jin, Peng and Yuan, Li},
  booktitle=emnlp,
  year={2024},
}

@inproceedings{zhulanguagebind,
  title={{LanguageBind}: Extending Video-Language Pretraining to N-modality by Language-based Semantic Alignment},
  author={Zhu, Bin and Lin, Bin and Ning, Munan and Yan, Yang and Cui, Jiaxi and Wang, HongFa and Pang, Yatian and Jiang, Wenhao and Zhang, Junwu and Li, Zongwei and Zhang, Cai and Li, Zhifeng and Liu, Wei and Yuan, Li},
  booktitle=iclr,
  year={2024},
}

@inproceedings{maaz2024video,
  title={{Video-ChatGPT}: Towards Detailed Video Understanding via Large Vision and Language Models},
  author={Maaz, Muhammad and Rasheed, Hanoona and Khan, Salman and Khan, Fahad Shahbaz},
  booktitle=acl,
  year={2024},
}

\appendix

\clearpage

\paragraph{Appendix overview.}
\begin{itemize}[nosep,leftmargin=1.5em]
  \item \Cref{app:setup} -- Experimental setup
  \item \Cref{app:validation} -- FSS and TempCore validation
  \item \Cref{app:ablations} -- Method ablations
  \item \Cref{app:detailed} -- Detailed results
\end{itemize}


\section{Experimental Setup}
\label{app:setup}

\subsection{Model Details}
\label{app:models}

We evaluate eight open-source VLMs spanning five model families
and three size tiers (256M--8B parameters).
The selection covers both established architectures (LLaVA-OneVision, InternVL)
and recent lightweight models (SmolVLM2, VideoLLaMA3),
ensuring diversity in design choices such as vision encoder integration,
temporal token handling, and training data composition.
All models are evaluated in a zero-shot, multiple-choice setting
with 32 input frames.

\begin{table}[!t]
\centering
\setlength{\tabcolsep}{4pt}
\resizebox{\columnwidth}{!}{%
\begin{tabular}{llr}
\toprule
Model & Architecture & Params \\
\midrule
LLaVA-OV-7B   & LLaVA-OneVision~\citep{li2024llava}     & 7B \\
LLaVA-OV-0.5B & LLaVA-OneVision~\citep{li2024llava}     & 0.5B \\
VideoLLaMA3-2B & VideoLLaMA3~\citep{zhang2025videollama}  & 2B \\
SmolVLM2-2.2B & SmolVLM2~\citep{marafioti2025smolvlm}   & 2.2B \\
SmolVLM2-500M & SmolVLM2~\citep{marafioti2025smolvlm}   & 500M \\
SmolVLM2-256M & SmolVLM2~\citep{marafioti2025smolvlm}   & 256M \\
Qwen3-VL-8B   & Qwen3-VL~\citep{qwen3vl_arxiv2025}     & 8B \\
InternVL3.5-8B & InternVL3~\citep{zhu2025internvl3}     & 8B \\
\bottomrule
\end{tabular}%
}
\caption{\textbf{Eight open-source VLMs spanning three size tiers (256M--8B).} Five model families evaluated in zero-shot, multiple-choice setting with 32 input frames.}
\label{tab:models}
\end{table}

\subsection{Benchmark Details}
\label{app:benchmarks}

We select six Video QA benchmarks that span a range of video lengths,
question types, and evaluation protocols.
Short-video benchmarks (EgoSchema, MVBench, NExTQA)
feature clips under three minutes and test fine-grained actions,
scene transitions, and causal reasoning.
Long-video benchmarks (MLVU, LongVideoBench, Video-MME)
contain videos ranging from several minutes up to an hour and require
multi-hop temporal reasoning over extended narratives.
All benchmarks use multiple-choice format,
enabling deterministic accuracy evaluation without LLM-based judging.

\textbf{MVBench}~\citep{li2024mvbench} assembles twenty
temporal task types by converting existing video annotations into multiple-choice
items, deliberately targeting skills a single frame should not satisfy.
\textbf{NExTQA}~\citep{xiao2021next} is organised around causal and temporal
questions over daily-life clips, asking why an action occurred or what followed it.
\textbf{EgoSchema}~\citep{mangalam2023egoschema} extends clips to three minutes of
egocentric footage and pairs them with narrative-rich options, so answering
requires integrating a long span rather than recognising one moment.
\textbf{MLVU}~\citep{mlvu_cvpr2025} spans nine task types across movies,
surveillance, egocentric video, cartoons and games, and is widely used to expose
degradation as length grows. \textbf{LongVideoBench}~\citep{wu2024longvideobench}
builds each question around a referring query pointing at a specific moment in
videos up to an hour long, which is why its questions are markedly
longer than the others'. \textbf{Video-MME}~\citep{fu2024videomme} is designed for
full-spectrum coverage, sampling short, medium and long videos across six domains.
The linguistic consequences of these choices are quantified in
\Cref{tab:benchmark_5axis_profile}.

\begin{table}[!t]
\centering
\setlength{\tabcolsep}{3pt}
\resizebox{\columnwidth}{!}{%
\begin{tabular}{lrrr}
\toprule
Benchmark & \# QA & Median Len. & Type \\
\midrule
EgoSchema~\citep{mangalam2023egoschema}     & 500   & $\sim$3 min  & Short \\
MVBench~\citep{li2024mvbench}              & 4{,}000 & $<$1 min   & Short \\
NExTQA~\citep{xiao2021next}               & 8{,}564 & $\sim$1 min  & Short \\
MLVU~\citep{mlvu_cvpr2025}                & 2{,}175 & $\sim$8 min  & Long \\
LongVideoBench~\citep{wu2024longvideobench} & 1{,}337 & $\sim$8 min  & Long \\
Video-MME~\citep{fu2024videomme}          & 2{,}700 & various      & Mixed \\
\bottomrule
\end{tabular}%
}
\caption{\textbf{Six Video QA benchmarks spanning seconds to tens of minutes.} ``Length'' is the median video duration; ``\# QA'' is the full evaluation set size.}
\label{tab:benchmarks}
\end{table}


\section{FSS and TempCore Validation}
\label{app:validation}

\subsection{Anti-key Samples as Temporally Sensitive}
\label{app:antikey_temporal}

Our main analysis includes Anti-key samples in \TempCore{}
alongside Key samples (\S\ref{sec:tempcore}).
The following evidence motivates this design decision:
\emph{Anti-key samples are also temporally sensitive}.
The key insight is that SigLIP's max-prob frames can be \emph{misleading}---selecting
visually matching but temporally wrong content
(\eg{}, for ``Did the yellow object leave the scene?'',
SigLIP selects frames \emph{with} the object rather than frames
showing its disappearance).

\begin{figure*}[t]
  \centering
  \includegraphics[width=\textwidth]{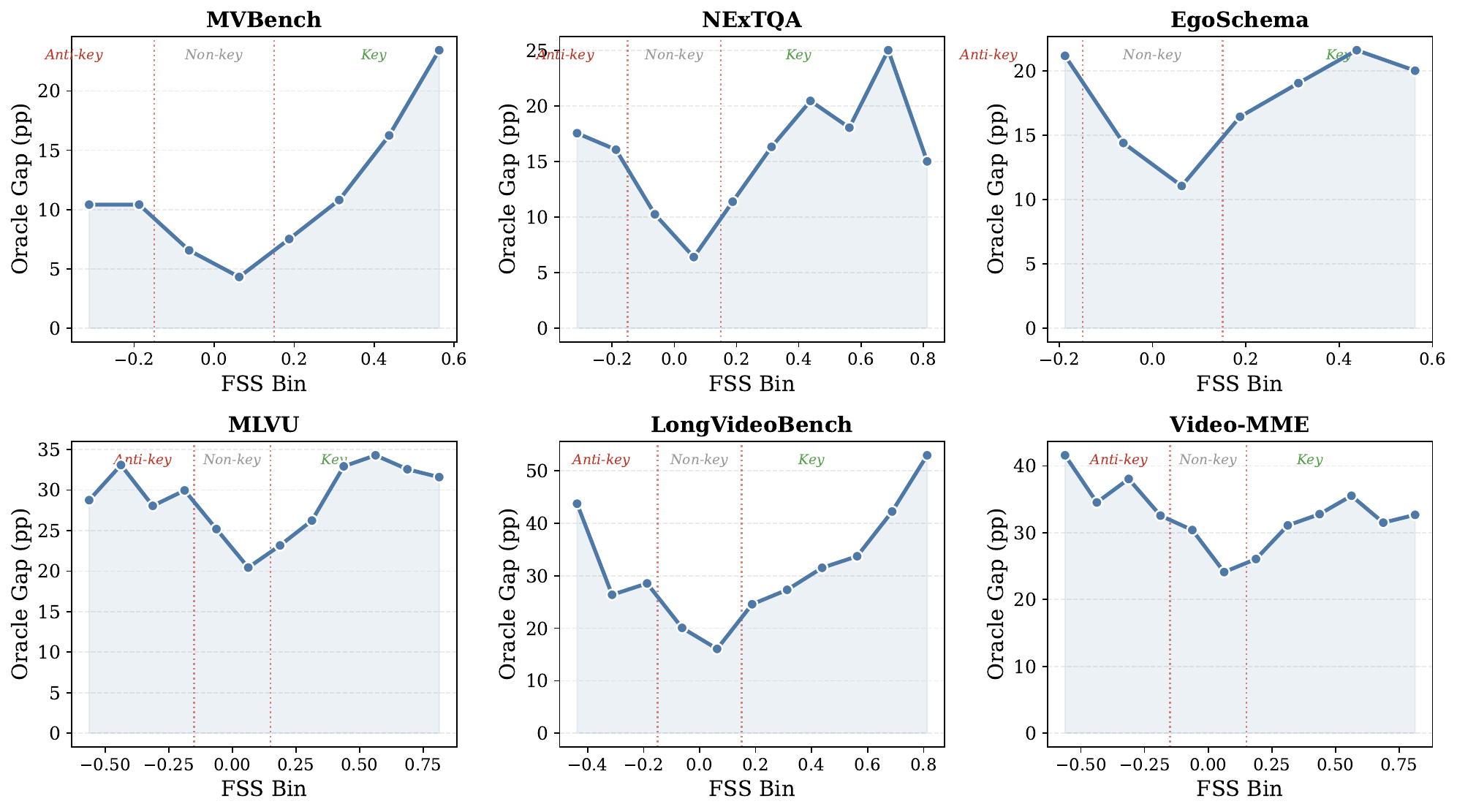}
  \caption{\textbf{Oracle gap exhibits a U-shape across FSS bins.}
  Both positive (Key) and negative (Anti-key) FSS tails show higher
  oracle gaps than the Non-key center, confirming that
  Anti-key samples are also temporally sensitive---they require
  specific temporal segments, but SigLIP selects misleading frames.
  Bins with fewer than 5 samples are omitted.}
  \label{fig:oracle_gap_fss_bin}
\end{figure*}

\begin{figure*}[t]
  \centering
  \includegraphics[width=\textwidth]{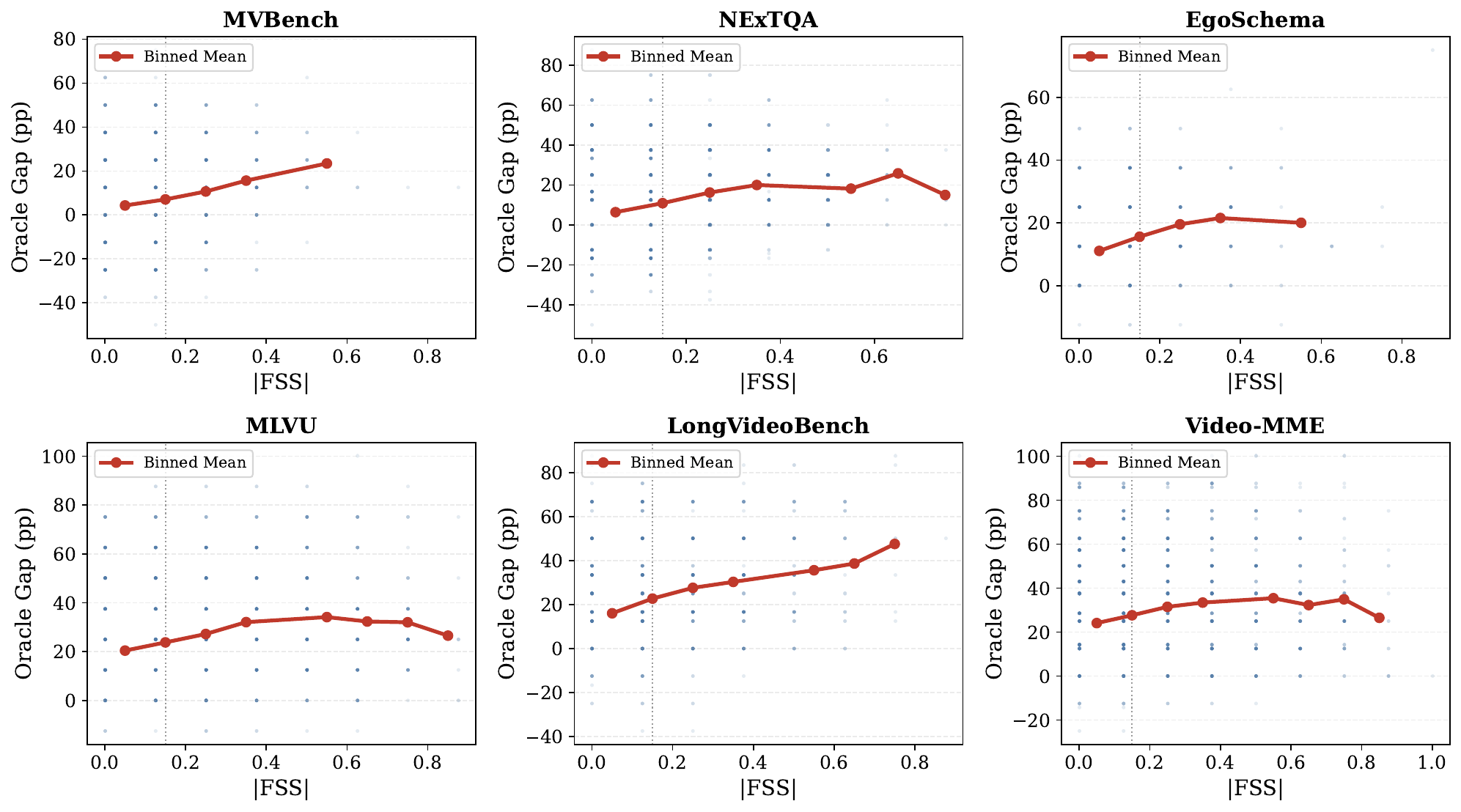}
  \caption{\textbf{Absolute FSS predicts oracle gap magnitude.}
  Per-sample scatter with binned means (red).
  The positive trend confirms that both high-FSS (Key) and
  low-FSS (Anti-key) samples benefit from temporal frame selection.}
  \label{fig:abs_fss_oracle}
\end{figure*}

We hypothesize two mechanisms underlying Anti-key behavior:
(i)~for \emph{static-appearance tasks} (\eg{}, Object Existence, Fine-grained Action),
SigLIP preferentially selects close-up shots of salient objects,
but the question asks about \emph{absence} or subtle state changes
that are better captured by contextual background frames;
(ii)~for \emph{temporal-sequence tasks} (\eg{}, Action Sequence),
the highest-similarity frames depict the endpoint of an action,
depriving the model of the preceding context needed to infer order.
In MLVU, the Anti-key fraction ($8.2\%$) is substantial
despite MLVU having the highest Key fraction ($36.0\%$),
indicating that visual appearance bias affects even temporally demanding benchmarks.

\paragraph{Oracle gap vs.\ FSS bin.}
\Cref{fig:oracle_gap_fss_bin} bins samples by quantized FSS values
and plots mean oracle gap per bin.
Both positive (Key) and negative (Anti-key) tails show elevated oracle gaps
compared to the near-zero center (Non-key),
producing a U-shaped curve.
This confirms that \emph{both} Key and Anti-key samples
benefit substantially from oracle frame selection,
while Non-key samples---which respond similarly regardless
of frame selection strategy---show lower oracle gaps.

\begin{table}[t]
\centering
\small
\setlength{\tabcolsep}{3pt}
\resizebox{\columnwidth}{!}{%
\begin{tabular}{llrrr|r}
\toprule
Benchmark & Category & Begin (\%) & Middle (\%) & End (\%) & Std \\
\midrule
\multicolumn{6}{l}{\textit{Short-video}} \\
MVBench & Key & 10.2 & 70.5 & 19.3 & 0.160 \\
 & Non-key & 5.3 & 85.8 & 8.9 & 0.119 \\
 & Anti-key & 0.0 & 92.7 & 7.3 & 0.090 \\
\midrule
NExTQA & Key & 7.0 & 83.0 & 10.0 & 0.127 \\
 & Non-key & 2.6 & 93.9 & 3.5 & 0.083 \\
 & Anti-key & 4.2 & 90.0 & 5.8 & 0.097 \\
\midrule
EgoSchema & Key & 10.6 & 76.6 & 12.8 & 0.135 \\
 & Non-key & 4.5 & 91.8 & 3.7 & 0.094 \\
 & Anti-key & 6.2 & 68.8 & 25.0 & 0.107 \\
\midrule
\midrule
\multicolumn{6}{l}{\textit{Long-video}} \\
MLVU & Key & 7.3 & 87.1 & 5.6 & 0.110 \\
 & Non-key & 2.4 & 95.9 & 1.7 & 0.078 \\
 & Anti-key & 1.7 & 92.7 & 5.6 & 0.096 \\
\midrule
LongVideoBench & Key & 5.0 & 90.3 & 4.7 & 0.105 \\
 & Non-key & 4.9 & 90.4 & 4.7 & 0.108 \\
 & Anti-key & 3.8 & 91.2 & 5.0 & 0.099 \\
\midrule
Video-MME & Key & 7.7 & 86.0 & 6.2 & 0.112 \\
 & Non-key & 4.0 & 92.3 & 3.8 & 0.093 \\
 & Anti-key & 2.9 & 93.6 & 3.5 & 0.085 \\
\bottomrule
\end{tabular}%
}
\caption{\textbf{Key samples' correct windows are spread across the video; Anti-key's are not consistently so.} Distribution of correct-window centroids across video thirds (Begin: $[0, 1/3)$, Middle: $[1/3, 2/3)$, End: $[2/3, 1]$). Std is the standard deviation of centroid position, so a higher value means correct windows sit at varied points in the video rather than defaulting to the middle. Key exceeds Non-key on five of six benchmarks (LongVideoBench is the exception); Anti-key exceeds it on three.}
\label{tab:centroid_position}
\end{table}

\paragraph{Oracle gap vs.\ $|\text{FSS}|$.}
\Cref{fig:abs_fss_oracle} plots per-sample oracle gap against $|\text{FSS}|$.
The binned mean (red line) rises overall:
higher $|\text{FSS}|$ correlates with larger oracle gaps
regardless of sign, reinforcing that both Key and Anti-key
samples are temporally localized.

\paragraph{Correct window centroid position.}
\Cref{tab:centroid_position} analyzes where in the video
the correct-answer windows are located.
Key samples show a higher centroid standard deviation than Non-key on five
of six benchmarks, so their correct windows sit at varied points in the video
rather than defaulting to the middle. Anti-key exceeds Non-key on three of six,
a weaker and less consistent effect.

\noindent
Together with the oracle gap subset analysis (\Cref{tab:oracle_subset}),
these results establish that Anti-key samples are temporally sensitive
but misled by visually plausible-but-temporally-wrong frames.
The true dichotomy is Temporal (Key + Anti-key) vs.\ Non-temporal (Non-key).

\subsection{Human Temporal Judgment Validation}
\label{app:human_eval}

To validate FSS against human perception, we conducted a preliminary human
evaluation study.
An annotator assessed stratified samples spanning all six benchmarks and the
three FSS categories (Key, Non-key, Anti-key at $\tau = 0.15$) on a 1--5
temporal necessity scale, passing where the necessity judgment felt unclear;
the rated set contains $N=100$ samples (32 Key, 48 Non-key, 20 Anti-key), with 9
further Non-key ratings excluded uniformly at random for balance. For each sample, annotators were shown the video, the question, and the answer options, and asked to assign a score indicating the extent to which temporal information was necessary to answer the question.

\paragraph{Metric correlations.}
\Cref{tab:human_metric_corr} reports individual metric correlations with
human temporal scores.
Crucially, $|\text{FSS}|$ (absolute value) achieves higher Spearman $\rho$
than signed FSS ($+0.391$ vs.\ $+0.376$),
confirming that \emph{both} Key and Anti-key samples exhibit temporal dependence.

\begin{table}[t]
\centering
\setlength{\tabcolsep}{5pt}
\begin{tabular}{lccc}
\toprule
Metric & Spearman $\rho$ & PB $r$ & AUC \\
\midrule
FSS & $+0.376$ & $+0.327$ & $0.685$ \\
$|\text{FSS}|$ & $+0.391$ & $+0.345$ & $0.726$ \\
\bottomrule
\end{tabular}
\caption{\textbf{Individual metric correlations with human temporal scores.} $|\text{FSS}|$ captures both Key and Anti-key temporal dependence, yielding higher correlation than signed FSS. Threshold for binarization: score $\geq$ 4. (PB $r$: Point-Biserial correlation.)}
\label{tab:human_metric_corr}
\end{table}

\paragraph{FSS category analysis.}
\Cref{tab:human_fss_bins} and \Cref{fig:human_fss_bin_boxplot} group human
temporal scores by FSS category.
Key samples score highest ($4.66$ mean, $96.9\%$ rated temporal), Anti-key
follow ($3.55$, $60.0\%$) and Non-key are lowest ($2.98$, $47.9\%$). The
Temporally Sensitive class (Key $\cup$ Anti-key) exceeds Non-key
($p < 0.001$, one-sided Mann--Whitney); Anti-key alone is directionally higher
but not significant at this sample size ($p = 0.18$).

\begin{table}[t]
\centering
\setlength{\tabcolsep}{4pt}
\begin{tabular}{lcccc}
\toprule
FSS Bin & $N$ & Mean & Temporal\% & High Conf.\% \\
\midrule
Key & 32 & 4.66 & 96.9\% & 100\% \\
Non-key & 48 & 2.98 & 47.9\% & 100\% \\
Anti-key & 20 & 3.55 & 60.0\% & 100\% \\
\bottomrule
\end{tabular}
\caption{\textbf{Human temporal scores by FSS category ($\tau=0.15$).} Key and Anti-key samples both exhibit elevated temporal scores compared to Non-key, supporting the U-shape hypothesis. Temporal\%: score $\geq$ 4.}
\label{tab:human_fss_bins}
\end{table}

\paragraph{$\tau$-sweep against human ground truth.}
\Cref{tab:tau_sweep_human} compares two temporal classification methods
at varying FSS thresholds, using human scores $\geq 4$ as ground truth.
Method~A (Key-only: FSS $> \tau$) achieves higher precision,
while Method~B (Key+Anti-key: $|\text{FSS}| > \tau$) gains in recall.
Method~B matches or exceeds Method~A at every $\tau$, confirming the
inclusion of Anti-key in \TempCore{}.
We adopt Method~B ($|\text{FSS}| > \tau$) as the default classification
throughout this work.

\begin{table}[t]
\centering
\setlength{\tabcolsep}{3pt}
\resizebox{\columnwidth}{!}{%
\begin{tabular}{c|ccc|ccc}
\toprule
 & \multicolumn{3}{c|}{Key-Only (FSS $> \tau$)} & \multicolumn{3}{c}{Key+Anti-Key ($|$FSS$| > \tau$)} \\
$\tau$ & Prec. & Rec. & F1 & Prec. & Rec. & F1 \\
\midrule
0.050 & 0.822 & 0.561 & 0.667 & 0.753 & 0.833 & \textbf{0.791} \\
0.100 & 0.822 & 0.561 & 0.667 & 0.753 & 0.833 & \textbf{0.791} \\
0.125 & 0.969 & 0.470 & 0.633 & 0.827 & 0.652 & \textbf{0.729} \\
0.150 & 0.969 & 0.470 & 0.633 & 0.827 & 0.652 & \textbf{0.729} \\
0.200 & 0.969 & 0.470 & 0.633 & 0.827 & 0.652 & \textbf{0.729} \\
0.250 & 1.000 & 0.303 & 0.465 & 0.852 & 0.348 & \textbf{0.495} \\
0.375 & 1.000 & 0.212 & 0.350 & 0.889 & 0.242 & \textbf{0.381} \\
0.500 & 1.000 & 0.136 & 0.240 & 0.909 & 0.152 & \textbf{0.260} \\
\bottomrule
\end{tabular}%
}
\caption{\textbf{$\tau$-sweep: Key-only vs.\ Key+Anti-key temporal classification against human ground truth.} Including Anti-key samples (right) improves recall and F1 across most $\tau$ values, confirming that anti-key frames also indicate temporal dependence.}
\label{tab:tau_sweep_human}
\end{table}

\subsection{External Correlation with Image VLM Benchmarks}
\label{app:external_corr}

To validate that \FSS{} correctly separates temporal from non-temporal performance,
we correlate Key-subset and Non-key-subset model rankings
with six image VLM benchmarks (MMMU, MMMU-Pro, MME, AI2D, TextVQA, OCRBench)
evaluated using \texttt{lmms-eval}~\citep{lmmseval2024}.
If \FSS{} correctly identifies frame-sensitive samples,
Non-key accuracy should correlate more strongly with image benchmarks
(measuring general VLM capability), while Key accuracy should diverge.

\Cref{tab:external_corr} shows the expected direction on every benchmark: Key
correlation never exceeds Non-key. The gap is widest on MLVU (mean
$\rho = 0.33$ against $0.78$) and Video-MME ($0.44$ against $0.78$), the two
benchmarks with the highest temporal purity (\Cref{tab:bqp}), and narrows to
$+0.26$ on MVBench and $+0.10$ on EgoSchema and LongVideoBench.

NExTQA is the exception: its Key and Non-key rankings coincide exactly
($\rho = 0.78$ for both), so the split carries no information about model
ranking there. With only eight models a Spearman correlation is coarse, and
NExTQA's Key set is small ($7.7\%$); we read the pattern as holding on the
benchmarks with substantial temporal content rather than universally.
This provides external evidence that \FSS{} captures
distinct, frame-sensitive capabilities rather than general VLM strength.

\begin{table}[h]
\centering
\setlength{\tabcolsep}{4pt}
\resizebox{\columnwidth}{!}{%
\begin{tabular}{lcccccc|c}
\toprule
& \multicolumn{6}{c|}{Spearman $\rho$ vs.\ Image VLM Benchmark} & \\
\cmidrule(lr){2-7}
Split & MMMU & MMMU-Pro & MME & AI2D & TextVQA & OCRBench & Mean \\
\midrule
\multicolumn{8}{l}{\textit{Short-video}} \\
\multicolumn{8}{l}{\textit{MVBench}} \\
\quad Key      & .64 & .64 & .62 & .55 & .57 & .31 & .56 \\
\quad Non-key  & .86 & .86 & .90 & .83 & .81 & .67 & .82 \\
\midrule
\multicolumn{8}{l}{\textit{NExTQA}} \\
\quad Key      & .83 & .83 & .86 & .79 & .76 & .60 & .78 \\
\quad Non-key  & .83 & .83 & .86 & .79 & .76 & .60 & .78 \\
\midrule
\multicolumn{8}{l}{\textit{EgoSchema}} \\
\quad Key      & .71 & .71 & .74 & .67 & .69 & .45 & .66 \\
\quad Non-key  & .81 & .81 & .83 & .76 & .71 & .64 & .76 \\
\midrule
\multicolumn{8}{l}{\textit{Long-video}} \\
\multicolumn{8}{l}{\textit{MLVU}} \\
\quad Key      & .43 & .43 & .40 & .31 & .33 & .10 & .33 \\
\quad Non-key  & .83 & .83 & .86 & .79 & .76 & .60 & .78 \\
\midrule
\multicolumn{8}{l}{\textit{LongVideoBench}} \\
\quad Key      & .79 & .79 & .83 & .76 & .74 & .55 & .74 \\
\quad Non-key  & .90 & .90 & .90 & .86 & .83 & .64 & .84 \\
\midrule
\multicolumn{8}{l}{\textit{Video-MME}} \\
\quad Key      & .52 & .52 & .57 & .45 & .43 & .17 & .44 \\
\quad Non-key  & .83 & .83 & .86 & .79 & .76 & .60 & .78 \\
\bottomrule
\end{tabular}%
}
\caption{\textbf{Non-key accuracy correlates more strongly with image VLM benchmarks than Key accuracy.} Spearman $\rho$ between Key and Non-key (including Anti-key) model rankings and six image benchmarks evaluated via \texttt{lmms-eval}. Key correlation is never higher than Non-key, and the gap is widest on MLVU and Video-MME, indicating that Key performance captures a distinct temporal skill rather than general VLM capability. $N\!=\!8$ models.}
\label{tab:external_corr}
\end{table}



\section{Method Ablations}
\label{app:ablations}

\subsection{Frame Count ($K$) Ablation}
\label{app:ablation}

We ablate the number of selected frames $K \in \{16, 32, 48\}$
on EgoSchema with LLaVA-OV-7B and LLaVA-OV-0.5B
(\Cref{tab:k_ablation}).
Accuracy differences across $K$ values are modest ($\leq$5\%),
suggesting that \maxprob{} is robust to the frame budget.
We use $K\!=\!32$ throughout as a practical default
that balances temporal coverage with computational cost.

\begin{table}[t]
\centering
\small
\resizebox{\columnwidth}{!}{%
\begin{tabular}{lrrr}
\toprule
Model & $K{=}16$ & $\boldsymbol{K{=}32}$ & $K{=}48$ \\
\midrule
LLaVA-OV-7B   & 50.6 & \textbf{53.8} & 55.0 \\
LLaVA-OV-0.5B & 23.8 & \textbf{24.2} & 25.0 \\
\bottomrule
\end{tabular}
}
\caption{\textbf{MaxProb accuracy is robust to frame budget $K$.} Q-mode accuracy (\%) on EgoSchema across $K \in \{16, 32, 48\}$ under the temporal-diversity selection of \S\ref{sec:method} for two representative models. Differences are modest, supporting $K{=}32$ as a practical default. \textbf{Bold} marks the default $K{=}32$.}
\label{tab:k_ablation}
\end{table}

\subsection{Benchmark Question Profiles}
\label{app:benchmark_profile}

Five-axis profile of the six benchmarks, referenced in \Cref{sec:setup}.
Descriptive density, WH-type entropy and type-token ratio are computed over the
full question set; ambiguity is a mean rating on a five-point scale.

\begin{table*}[t]
\centering
\small
\setlength{\tabcolsep}{16pt}
\resizebox{\textwidth}{!}{%
\begin{tabular}{lrrrrrr}
\toprule
Benchmark & N & Q-words & Desc.\% & WH-ent & TTR & Ambig. \\
\midrule
MVBench & 4,000 & 13.1 & 1.81 & 2.04 & 0.038 & 1.63 \\
NExTQA & \textbf{8,564} & 11.5 & 3.93 & 1.84 & 0.028 & 1.61 \\
EgoSchema & 500 & 24.5 & 0.14 & 1.24 & 0.114 & \textbf{2.71} \\
MLVU & 2,174 & 15.4 & 1.97 & \textbf{2.32} & 0.071 & 1.29 \\
LongVideoBench & 1,337 & \textbf{44.1} & \textbf{8.47} & 0.71 & 0.104 & 1.33 \\
Video-MME & 2,700 & 13.6 & 1.25 & 2.27 & \textbf{0.151} & 1.47 \\
\bottomrule
\end{tabular}%
}
\caption{\textbf{The six benchmarks differ on every axis we can measure, so median length alone does not characterise them.} ``N'': evaluation-set size; ``Q-words'': mean question length. ``Desc.\%'': share of question tokens that are visual descriptors (colour, shape, object). ``WH-ent'': Shannon entropy in bits over 11 WH-type bins, so higher means WH-types are evenly mixed and lower means one type dominates. ``TTR'': type-token ratio over question text, a lexical-diversity index. ``Ambig.'': mean of two independent LLM 1--5 under-specification scores (1 = clear). Column maxima in bold.}
\label{tab:benchmark_5axis_profile}
\end{table*}

\subsection{Frame Coverage and Sensitivity}
\label{app:kcoverage}

\Cref{tab:kcoverage_binning} reports mean $|\FSS{}|$ binned by frame coverage
$K/L$, the ratio of the frame budget to the number of scored frames. Only MVBench reaches the highest coverage bin;
EgoSchema and MLVU sit entirely below $K/L = 0.10$. ``--'' marks an empty bin.

\begin{table}[t]
\centering
\small
\setlength{\tabcolsep}{4pt}
\resizebox{\columnwidth}{!}{%
\begin{tabular}{lrrrr}
\toprule
Benchmark & $\geq$0.5 & 0.25--0.5 & 0.10--0.25 & $<$0.10 \\
\midrule
MVBench & 4.5 & 7.2 & 7.0 & 9.5 \\
NExTQA & -- & 5.6 & 6.0 & 8.3 \\
EgoSchema & -- & -- & -- & 11.9 \\
MLVU & -- & -- & -- & 20.9 \\
LongVideoBench & -- & 6.1 & 6.6 & 15.8 \\
Video-MME & -- & 2.1 & 7.8 & 18.4 \\
\bottomrule
\end{tabular}%
}
\caption{\textbf{The scarcest frame budget yields the largest $|$FSS$|$ in every benchmark.} Cell: 8-model average $|$FSS$|$ (\%) for samples binned by K/L, the ratio of the K=32 frame budget to the number of scored frames $L$. The $<$0.10 bin holds the row maximum everywhere, exceeding the best-covered bin by 2.7--16.3\,pp in the four benchmarks that populate more than one bin. Frame selection therefore matters in proportion to how much of the video the budget must discard. ``--'' indicates no samples in that bin.}
\label{tab:kcoverage_binning}
\end{table}

\subsection{Contamination Probe}
\label{app:leakage}

Two controls test whether no-frame (\LIS{}) accuracy reflects benchmark
contamination acquired during video-multimodal training. First, we extend the
\LIS{} model panel with three text-only LLMs, for which video-specific
contamination is impossible by construction (\Cref{tab:lis_llm});
benchmark-level \LIS{} moves by at most $2.9$\,pp and by under $1$\,pp on three
of six. These probes cover a $300$-sample subset per benchmark, so this is a
robustness check on the aggregate, and per-sample \LIS{} throughout the paper
remains the 8-VLM quantity.

Second, a same-backbone paired control (\Cref{tab:paired_backbone}) compares
Qwen3-8B against Qwen3-VL-8B-Instruct, which adds video-multimodal post-training
on top of it. Both see the same questions with no usable visual input, so the
difference bounds what that post-training contributes, contamination included.
On MVBench, NExTQA, EgoSchema and MLVU it is $+1.0$ to $+2.3$\,pp. On
LongVideoBench ($+13.3$) and Video-MME ($+8.7$) it is not small, and our probes
cannot separate benchmark exposure from ordinary gains in multiple-choice
instruction following; the Trivial fractions of those two benchmarks should be
read with that reservation.

\begin{table}[t]
\centering
\small
\setlength{\tabcolsep}{4pt}
\resizebox{\columnwidth}{!}{%
\begin{tabular}{lrrr}
\toprule
Benchmark & 8 VLMs & $+$ 3 LLMs & $\Delta$ (pp) \\
\midrule
MVBench        & 29.1 & 30.8 & +1.6 \\
NExTQA         & 34.2 & 37.1 & +2.9 \\
EgoSchema      & 25.7 & 27.5 & +1.8 \\
MLVU           & 36.1 & 36.3 & +0.2 \\
LongVideoBench & 35.3 & 35.1 & -0.2 \\
Video-MME      & 33.1 & 33.2 & +0.0 \\
\bottomrule
\end{tabular}%
}
\caption{\textbf{Adding models that have never seen a video barely moves \LIS{}.} Mean no-frame accuracy (\%) over the 8-VLM panel versus the same panel plus 3 text-only LLMs (Qwen2.5-7B, Llama-3.1-8B, Qwen3-8B), for which video-specific contamination is impossible by construction. The largest shift is $+2.9$\,pp and three of six benchmarks move by under $1$\,pp, so the language-prior signal \LIS{} measures is not an artefact of video-multimodal training. The LLM probes cover a $300$-sample subset per benchmark, so this is a benchmark-level robustness check; per-sample \LIS{} throughout the paper remains the 8-VLM quantity.}
\label{tab:lis_llm}
\end{table}

\begin{table}[t]
\centering
\small
\setlength{\tabcolsep}{4pt}
\resizebox{\columnwidth}{!}{%
\begin{tabular}{lrrrr}
\toprule
Benchmark & $N$ & Qwen3-8B & Qwen3-VL-8B & $\Delta$ \\
 &  & (text-only) & (no-frame) & \\
\midrule
MVBench        & 300 & 37.3 & 39.3 & +2.0 \\
NExTQA         & 300 & 46.7 & 49.0 & +2.3 \\
EgoSchema      & 300 & 33.7 & 35.7 & +2.0 \\
MLVU           & 300 & 38.7 & 39.7 & +1.0 \\
LongVideoBench & 300 & 32.0 & 45.3 & +13.3 \\
Video-MME      & 300 & 35.3 & 44.0 & +8.7 \\
\midrule
Mean & & & & +4.9 \\
\bottomrule
\end{tabular}%
}
\caption{\textbf{Video post-training leaves no-frame accuracy nearly unchanged on four benchmarks but adds substantially on two.} Qwen3-8B is a text-only LLM; Qwen3-VL-8B-Instruct shares that backbone and adds video-multimodal post-training. Both answer the same questions with no usable visual input, so $\Delta$ upper-bounds what that post-training contributes---including any benchmark exposure it carried, but also ordinary gains in multiple-choice instruction following. Columns are computed on identical samples; every probe item was checked to carry the dataset's own gold letter.}
\label{tab:paired_backbone}
\end{table}

\subsection{Random and Uniform Baselines}
\label{app:random_baseline}

\begin{table}[t]
\centering
\small
\setlength{\tabcolsep}{4pt}
\resizebox{\columnwidth}{!}{%
\begin{tabular}{lrrrrr}
\toprule
Benchmark & MinProb & Random & Uniform & MaxProb & Max$-$Random \\
\midrule
MVBench & 47.4 & 48.7 & 48.7 & 49.0 & +0.3 \\
NExTQA & 56.8 & 58.4 & 59.4 & 59.2 & +0.8 \\
EgoSchema & 32.7 & 37.8 & 38.1 & 39.2 & +1.4 \\
MLVU & 44.1 & 50.9 & 51.5 & 57.3 & +6.4 \\
LongVideoBench & 41.4 & 44.3 & 45.2 & 47.8 & +3.5 \\
Video-MME & 38.6 & 46.1 & 46.3 & 49.7 & +3.7 \\
\bottomrule
\end{tabular}%
}
\caption{\textbf{Random sampling baseline confirms the encoder-based selection signal.} Mean accuracy across 8 VLMs (\%), QA-mode scoring with $K=32$ frames, on the samples with complete results under all four strategies. MaxProb exceeds Random on all 6 benchmarks (largest margin +6.4\,pp on MLVU) and Random exceeds MinProb on 6 of 6, so the gap reflects ranking quality rather than a sampling-protocol artifact. ``Uniform'' samples $K$ frames at equal intervals; it tracks Random closely and edges MaxProb only on NExTQA ($59.4$ vs.\ $59.2$).}
\label{tab:random_baseline}
\end{table}

\subsection{Per-Encoder Comparison}
\label{app:encoder}

We compare \maxprob{} accuracy with three vision-language encoders:
SigLIP2-400M (default), SigLIP-400M, and CLIP ViT-L/14.
\Cref{tab:encoder_comparison} reports Q-mode accuracy
on the Perception benchmark (19{,}140 samples) for two LLaVA models.
All three encoders produce nearly identical accuracy,
with differences within 0.5\,pp.
This confirms that frame selection quality
is robust to the choice of scoring encoder,
consistent with the finding of \citet{apollo_cvpr2025}
that modern vision-language encoders converge in frame-level ranking ability.

\begin{table}[h]
\centering
\setlength{\tabcolsep}{5pt}
\resizebox{\columnwidth}{!}{%
\begin{tabular}{lrrr}
\toprule
Model & SigLIP2-400M & SigLIP-400M & CLIP ViT-L \\
\midrule
LLaVA-OV-7B  & 57.5 & 57.5 & 57.9 \\
LLaVA-OV-0.5B & 48.8 & 48.8 & 48.9 \\
\midrule
$\Delta$ (max$-$min) & \multicolumn{3}{c}{$\leq 0.5$\,pp} \\
\bottomrule
\end{tabular}%
}
\caption{\textbf{Encoder choice has negligible impact on MaxProb accuracy.}
Q-mode MaxProb accuracy (\%) on Perception (19{,}140 samples) with three vision-language encoders.
All differences are within 0.5\,pp, confirming that frame selection quality is robust to encoder choice.}
\label{tab:encoder_comparison}
\end{table}

\subsection{Question and Answer Text Statistics}
\label{app:text_stats}

\Cref{tab:text_stats} reports the mean question and answer option lengths
across six benchmarks.
LongVideoBench questions average 44.1 words,
up to nearly four times longer than those of other benchmarks (12--25 words),
because they embed detailed scene descriptions
---clothing, spatial layout, character actions---before posing a query.
This explains why QA-mode adds least on LongVideoBench among the long-video benchmarks
(\S\ref{ssec:rq1}): the question text alone provides
sufficient visual grounding for the encoder,
and appending answer text adds noise rather than signal.

\begin{table}[h]
\centering
\setlength{\tabcolsep}{5pt}
\resizebox{\columnwidth}{!}{%
\begin{tabular}{lrrr}
\toprule
Benchmark & Q (words) & A (words) & \#Opts \\
\midrule
MVBench        & 13.1 & 4.0  & 3.7 \\
NExTQA         & 11.5 & 2.8  & 5.0 \\
EgoSchema      & 24.5 & 22.5 & 5.0 \\
MLVU           & 15.4 & 2.5  & 4.0 \\
\textbf{LongVideoBench} & \textbf{44.1} & \textbf{9.0} & \textbf{4.7} \\
Video-MME      & 13.6 & 5.1  & 4.0 \\
\bottomrule
\end{tabular}%
}
\caption{\textbf{LongVideoBench questions are up to nearly $4\times$ longer than other benchmarks'.} Mean question length (Q), answer option length (A), and number of options (\#Opts) across six benchmarks. The descriptive nature of LongVideoBench questions explains why QA-mode adds least there among the long-video benchmarks.}
\label{tab:text_stats}
\end{table}

\paragraph{LongVideoBench case study.}
The following examples illustrate the descriptive nature
of LongVideoBench questions:

\begin{quote}
\small
\textbf{LongVideoBench (77 words):}
``In a room decorated with various paintings,
there are two men standing in the middle.
On the left is a man with short hair wearing a red short-sleeve shirt, holding a beer.
On the right is a man with a black beard, wearing a gray short-sleeve shirt.
Behind the man in red is a white door.
When the subtitle mentions `over 2600 meters yeah there you go we',
what did the man in the red shirt do?''
\end{quote}

\begin{quote}
\small
\textbf{LongVideoBench (79 words):}
``In the video, there are two men in a room.
The man on the left is wearing a yellow shirt, a hat backwards,
and a watch on his left hand.
The man on the right is wearing a shirt with red flowers and green leaves
and has a bracelet on his left hand.
When the subtitle mentions `interested enough to join us but first,'
what change happens to the man wearing the shirt
with red flowers and green leaves?''
\end{quote}

\begin{quote}
\small
\textbf{MLVU (4 words):}
``Did I drink water?''
\end{quote}

\noindent
The contrast is striking: LongVideoBench questions specify
clothing colors, accessories, and spatial positions in enough detail
for a vision-language encoder to localize the relevant frames
from the question alone.
In contrast, short questions like the MLVU example
provide almost no visual cues, making the answer text
a valuable source of grounding signal.


\section{Detailed Results}
\label{app:detailed}

\subsection{Per-Benchmark Figures}
\label{app:fss_detail}

\begin{figure*}[t]
  \centering
  \includegraphics[width=\textwidth]{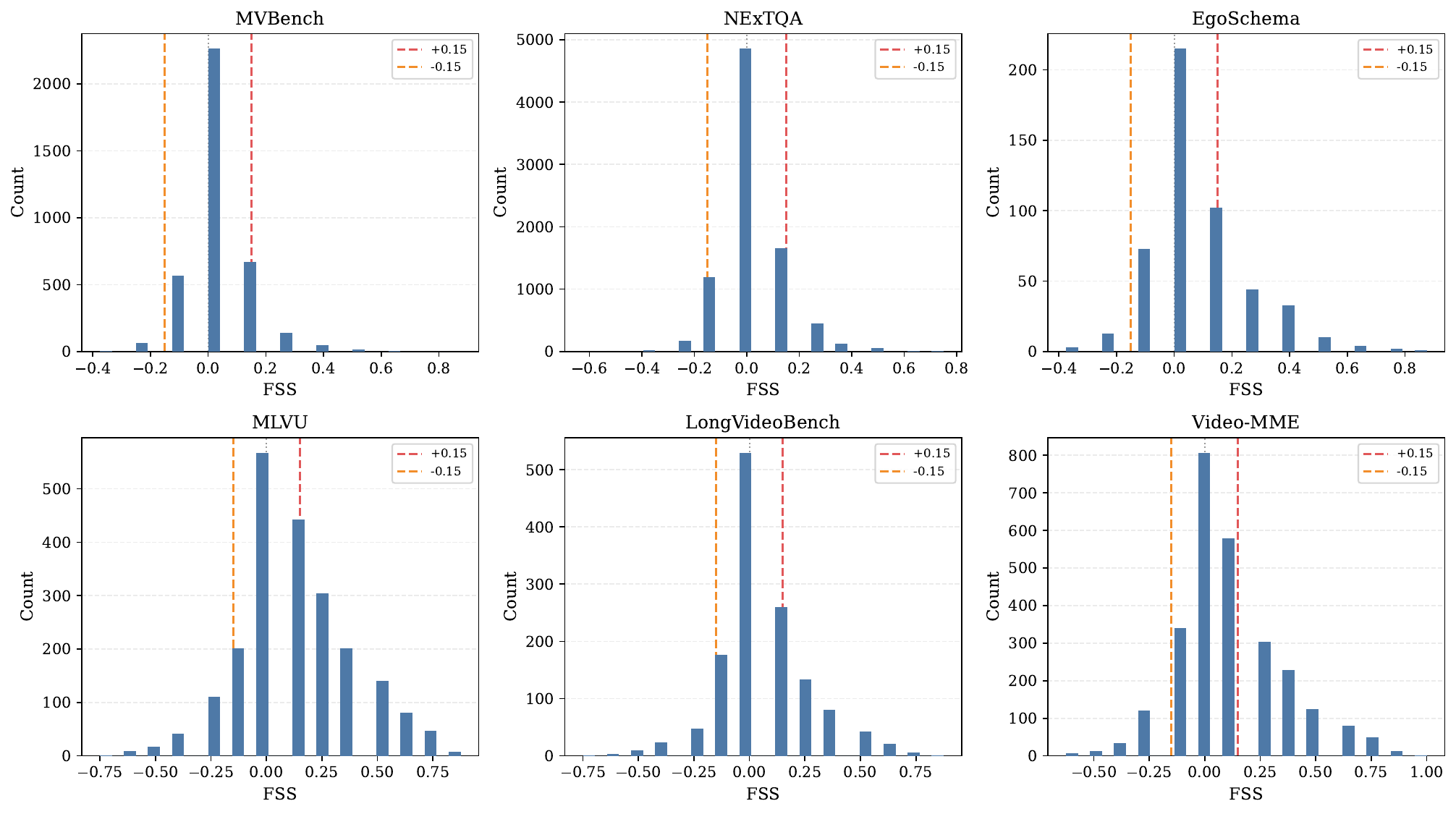}
  \caption{\textbf{Per-benchmark \FSS{} distributions ($\tau = 0.15$).} All six benchmarks show a dominant near-zero peak, with MLVU exhibiting the broadest spread.}
  \label{fig:fss_dist_detail}
\end{figure*}

\begin{figure*}[t]
  \centering
  \includegraphics[width=\textwidth]{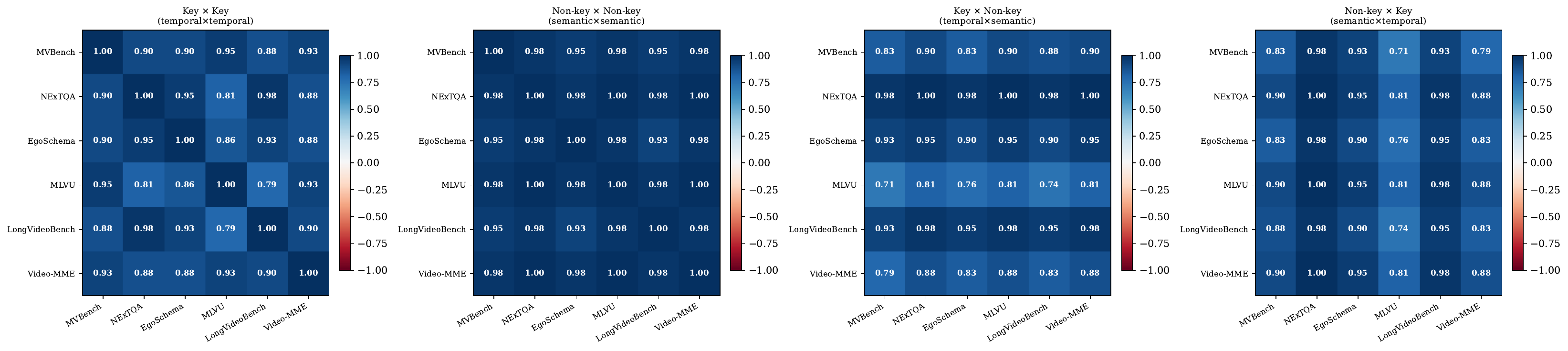}
  \caption{\textbf{Cross-benchmark model ranking correlations.} $1 \times 4$ Spearman $\rho$ matrices for Key$\times$Key, Non-key$\times$Non-key, Key$\times$Non-key, and Non-key$\times$Key subsets.}
  \label{fig:fss_corr_detail}
\end{figure*}

\begin{figure*}[t]
  \centering
  \includegraphics[width=\textwidth]{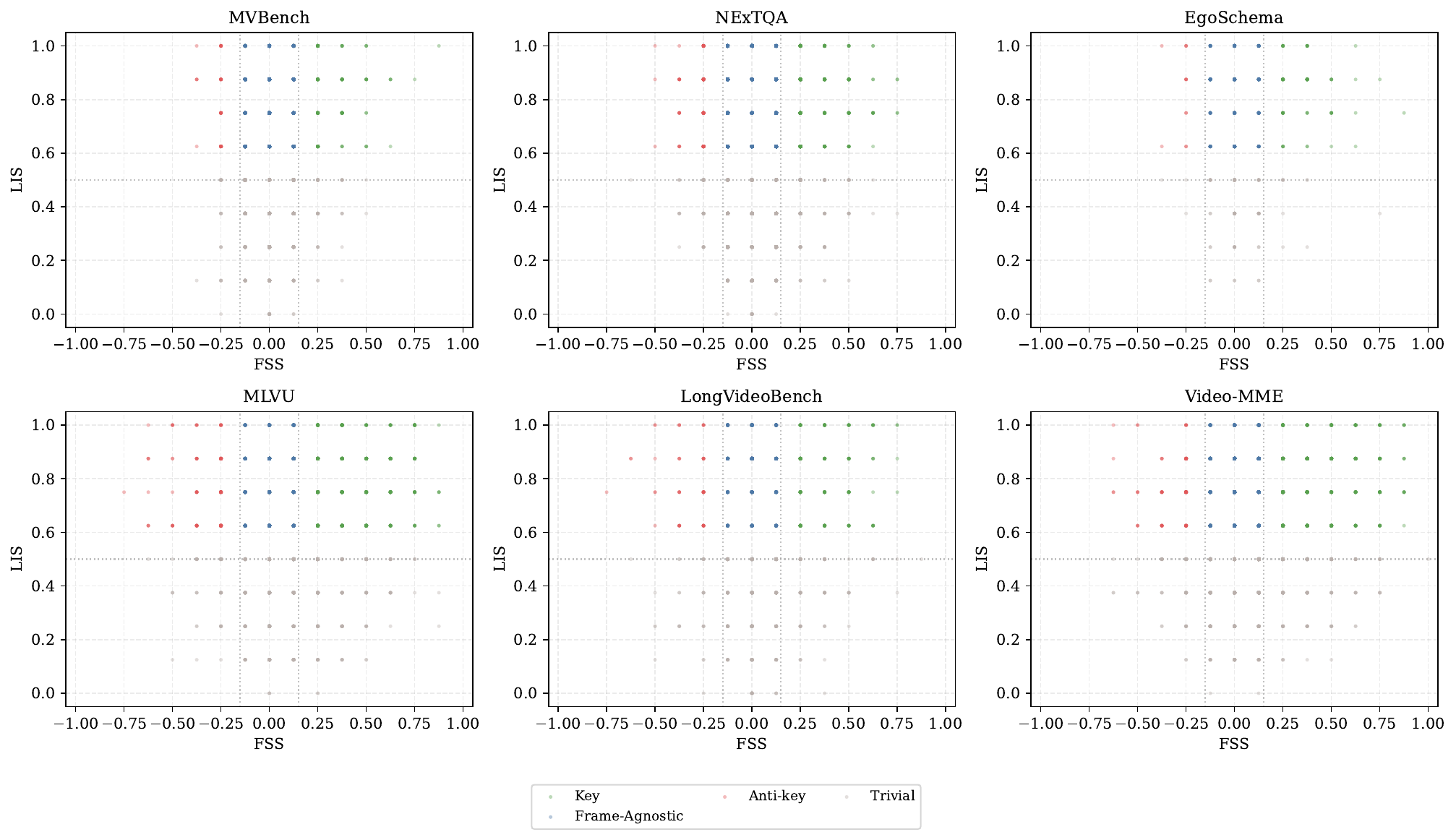}
  \caption{\textbf{Per-benchmark LIS--FSS scatter plots.} Each panel shows the two-dimensional (LIS, FSS) distribution with four-category coloring. Dashed lines mark thresholds ($\tau = 0.15$, $\lambda = 0.50$).}
  \label{fig:lis_fss_detail}
\end{figure*}

\Cref{fig:fss_dist_detail} presents the per-benchmark \FSS{} distributions
corresponding to the aggregated view in \Cref{fig:fss_dist}.
\Cref{fig:fss_corr_detail} shows the cross-benchmark model ranking correlation matrices;
\Cref{fig:lis_fss_detail} shows the per-benchmark LIS--FSS scatter plots.

\subsection{Performance Tables}
\label{app:perf_detail}

\Cref{tab:mode_comparison_full}--\Cref{tab:cross_strategy_videomme} provide
complete per-model results for each evaluation strategy and benchmark.

\begin{table*}[t]
\centering
\small
\caption{\textbf{QA-mode scoring wins in 42 of 48 model$\times$benchmark cells.} Per-model Q-mode vs.\ QA-mode \maxprob{} accuracy (\%) across all six benchmarks; \Cref{tab:mode_comparison} reports the model-averaged summary. Five of the six exceptions are VideoLLaMA3-2B, and none exceeds $1.4$\,pp, so the advantage is broad rather than driven by a few models.}
\label{tab:mode_comparison_full}
\resizebox{\textwidth}{!}{%
\begin{tabular}{lrrrrrrrrrrrr}
\toprule
 & \multicolumn{6}{c}{\textit{Short-video}} & \multicolumn{6}{c}{\textit{Long-video}} \\
\cmidrule(lr){2-7}
\cmidrule(lr){8-13}
 & \multicolumn{2}{c}{MVBench} & \multicolumn{2}{c}{NExTQA} & \multicolumn{2}{c}{EgoSchema} & \multicolumn{2}{c}{MLVU} & \multicolumn{2}{c}{LongVideoBench} & \multicolumn{2}{c}{Video-MME} \\
\cmidrule(lr){2-3}
\cmidrule(lr){4-5}
\cmidrule(lr){6-7}
\cmidrule(lr){8-9}
\cmidrule(lr){10-11}
\cmidrule(lr){12-13}
Model & Q & QA & Q & QA & Q & QA & Q & QA & Q & QA & Q & QA \\
\midrule
InternVL3.5-8B & \textbf{68.7} & \textbf{69.5} & \textbf{79.5} & \textbf{80.7} & \textbf{55.2} & \textbf{66.2} & \textbf{68.7} & \textbf{72.6} & \textbf{59.2} & \textbf{61.1} & \textbf{59.3} & \textbf{62.8} \\
LLaVA-OV-7B & 55.8 & 57.0 & 78.0 & 78.7 & 53.8 & 62.8 & 67.6 & 70.8 & 58.3 & 59.3 & 56.8 & 62.2 \\
Qwen3-VL-8B & 56.4 & 57.6 & 70.9 & 71.2 & 50.6 & 58.2 & 57.4 & 60.1 & 50.2 & 52.4 & 50.5 & 54.1 \\
SmolVLM2-2.2B & 46.5 & 47.4 & 60.8 & 62.3 & 25.8 & 32.0 & 58.0 & 59.9 & 50.2 & 50.6 & 47.5 & 53.2 \\
LLaVA-OV-0.5B & 46.3 & 47.4 & 57.2 & 58.5 & 24.2 & 27.4 & 57.7 & 59.9 & 45.3 & 46.7 & 44.5 & 49.3 \\
SmolVLM2-500M & 42.0 & 42.3 & 46.9 & 48.3 & 23.4 & 26.6 & 51.4 & 55.2 & 42.5 & 43.8 & 40.5 & 45.7 \\
VidLLaMA3-2B & 40.2 & 38.8 & 46.9 & 46.4 & 23.2 & 23.0 & 42.9 & 42.8 & 41.1 & 40.9 & 36.7 & 36.8 \\
SmolVLM2-256M & 31.6 & 31.9 & 27.6 & 27.4 & 17.0 & 17.2 & 35.6 & 37.4 & 26.8 & 27.7 & 32.5 & 33.7 \\
\bottomrule
\end{tabular}%
}
\end{table*}

\begin{table*}[t]
\centering
\resizebox{\textwidth}{!}{%
\begin{tabular}{lrrrrrrrrrrrrrrrrrrrr}
\toprule
 & \multicolumn{9}{c}{\textit{Short-video}} & \multicolumn{9}{c}{\textit{Long-video}} & \multicolumn{2}{c}{\textit{Avg.\ $\Delta$}} \\
\cmidrule(lr){2-10}
\cmidrule(lr){11-19}
\cmidrule(lr){20-21}
 & \multicolumn{3}{c}{MVBench} & \multicolumn{3}{c}{NExTQA} & \multicolumn{3}{c}{EgoSchema} & \multicolumn{3}{c}{MLVU} & \multicolumn{3}{c}{LongVideoBench} & \multicolumn{3}{c}{Video-MME} &  &  \\
\cmidrule(lr){2-4}
\cmidrule(lr){5-7}
\cmidrule(lr){8-10}
\cmidrule(lr){11-13}
\cmidrule(lr){14-16}
\cmidrule(lr){17-19}
Model & Max & Min & $\Delta$ & Max & Min & $\Delta$ & Max & Min & $\Delta$ & Max & Min & $\Delta$ & Max & Min & $\Delta$ & Max & Min & $\Delta$ & Short & Long \\
\midrule
InternVL3.5-8B & \textbf{69.5} & \textbf{67.2} & +2.3 & \textbf{80.7} & \textbf{77.0} & +3.6 & \textbf{66.2} & \textbf{53.2} & +13.0 & \textbf{72.6} & \textbf{53.4} & +19.2 & \textbf{61.1} & \textbf{50.3} & +10.8 & \textbf{62.8} & \textbf{47.9} & +15.0 & +6.3 & +15.0 \\
LLaVA-OV-7B & 57.0 & 54.8 & +2.2 & 78.7 & 75.0 & +3.7 & 62.8 & 49.0 & +13.8 & 70.8 & 49.9 & +20.9 & 59.3 & 47.9 & +11.4 & 62.2 & 43.4 & +18.8 & +6.6 & +17.0 \\
Qwen3-VL-8B & 57.6 & 56.2 & +1.4 & 71.2 & 68.9 & +2.3 & 58.2 & 50.2 & +8.0 & 60.1 & 48.6 & +11.5 & 52.4 & 46.1 & +6.4 & 54.1 & 44.5 & +9.6 & +3.9 & +9.1 \\
SmolVLM2-2.2B & 47.4 & 45.5 & +2.0 & 62.3 & 58.8 & +3.5 & 32.0 & 24.8 & +7.2 & 59.9 & 45.5 & +14.4 & 50.6 & 43.0 & +7.6 & 53.2 & 39.1 & +14.1 & +4.2 & +12.0 \\
LLaVA-OV-0.5B & 47.4 & 44.4 & +3.0 & 58.5 & 54.7 & +3.8 & 27.4 & 21.4 & +6.0 & 59.9 & 43.4 & +16.6 & 46.7 & 38.6 & +8.1 & 49.3 & 34.7 & +14.6 & +4.2 & +13.1 \\
SmolVLM2-500M & 42.3 & 40.7 & +1.6 & 48.3 & 45.6 & +2.6 & 26.6 & 23.2 & +3.4 & 55.2 & 41.0 & +14.2 & 43.8 & 38.1 & +5.7 & 45.7 & 33.9 & +11.8 & +2.5 & +10.6 \\
VidLLaMA3-2B & 38.8 & 39.1 & -0.2 & 46.4 & 46.8 & -0.4 & 23.0 & 23.4 & -0.4 & 42.8 & 41.8 & +1.0 & 40.9 & 40.5 & +0.4 & 36.8 & 37.4 & -0.6 & -0.4 & +0.3 \\
SmolVLM2-256M & 31.9 & 31.4 & +0.5 & 27.4 & 27.1 & +0.3 & 17.2 & 16.2 & +1.0 & 37.4 & 29.6 & +7.8 & 27.7 & 27.1 & +0.7 & 33.7 & 27.7 & +6.0 & +0.6 & +4.8 \\
\midrule
\textit{Average} & 49.0 & 47.4 & +1.6 & 59.2 & 56.8 & +2.4 & 39.2 & 32.7 & +6.5 & 57.3 & 44.1 & +13.2 & 47.8 & 41.4 & +6.4 & 49.7 & 38.6 & +11.2 & +3.5 & +10.2 \\
\bottomrule
\end{tabular}%
}
\caption{\textbf{MaxProb outperforms MinProb on long-video benchmarks in 23 of 24 model--benchmark cells.} Per-model accuracy (\%) with Short and Long average $\Delta$ columns. Best value per column in bold.}
\label{tab:maxmin}
\end{table*}

\begin{table*}[t]
\centering
\resizebox{\textwidth}{!}{%
\begin{tabular}{lrrrrrrrrrrrrrrrrrrrr}
\toprule
 & \multicolumn{9}{c}{\textit{Short-video}} & \multicolumn{9}{c}{\textit{Long-video}} & \multicolumn{2}{c}{\textit{Avg.\ Gap}} \\
\cmidrule(lr){2-10}
\cmidrule(lr){11-19}
\cmidrule(lr){20-21}
 & \multicolumn{3}{c}{MVBench} & \multicolumn{3}{c}{NExTQA} & \multicolumn{3}{c}{EgoSchema} & \multicolumn{3}{c}{MLVU} & \multicolumn{3}{c}{LongVideoBench} & \multicolumn{3}{c}{Video-MME} &  &  \\
\cmidrule(lr){2-4}
\cmidrule(lr){5-7}
\cmidrule(lr){8-10}
\cmidrule(lr){11-13}
\cmidrule(lr){14-16}
\cmidrule(lr){17-19}
Model & Uniform & Oracle & Gap & Uniform & Oracle & Gap & Uniform & Oracle & Gap & Uniform & Oracle & Gap & Uniform & Oracle & Gap & Uniform & Oracle & Gap & Short & Long \\
\midrule
InternVL3.5-8B & \textbf{69.1} & \textbf{76.7} & +7.5 & \textbf{80.2} & 87.7 & +7.4 & \textbf{64.6} & \textbf{83.8} & +19.2 & \textbf{65.6} & \textbf{93.7} & +28.1 & \textbf{57.3} & \textbf{83.9} & +26.6 & \textbf{61.3} & 85.2 & +24.0 & +11.4 & +26.2 \\
LLaVA-OV-7B & 57.0 & 65.1 & +8.1 & 78.1 & \textbf{88.2} & +10.0 & 63.0 & 83.4 & +20.4 & 63.9 & 92.7 & +28.8 & 56.5 & 81.8 & +25.3 & 57.7 & \textbf{87.9} & +30.1 & +12.8 & +28.1 \\
Qwen3-VL-8B & 57.2 & 66.4 & +9.2 & 70.5 & 84.4 & +13.9 & 54.0 & 81.0 & +27.0 & 52.0 & 87.3 & +35.3 & 48.8 & 80.9 & +32.1 & 49.7 & 85.6 & +36.0 & +16.7 & +34.5 \\
LLaVA-OV-0.5B & 46.4 & 52.1 & +5.7 & 57.0 & 69.9 & +13.0 & 26.4 & 40.2 & +13.8 & 51.3 & 82.8 & +31.5 & 43.0 & 63.8 & +20.8 & 44.3 & 75.0 & +30.7 & +10.8 & +27.7 \\
SmolVLM2-2.2B & 46.7 & 52.1 & +5.4 & 61.1 & 70.4 & +9.3 & 29.4 & 41.8 & +12.4 & 54.6 & 77.0 & +22.5 & 47.3 & 67.8 & +20.6 & 49.4 & 73.8 & +24.4 & +9.1 & +22.5 \\
SmolVLM2-500M & 42.1 & 47.7 & +5.6 & 47.2 & 57.4 & +10.3 & 26.2 & 44.4 & +18.2 & 48.6 & 78.5 & +30.0 & 40.9 & 69.2 & +28.3 & 40.3 & 75.1 & +34.8 & +11.4 & +31.0 \\
SmolVLM2-256M & 31.5 & 33.6 & +2.1 & 27.9 & 37.2 & +9.3 & 17.0 & 22.2 & +5.2 & 33.3 & 60.4 & +27.1 & 26.8 & 48.9 & +22.1 & 30.8 & 67.5 & +36.7 & +5.5 & +28.6 \\
VidLLaMA3-2B & 39.5 & 41.0 & +1.5 & 47.0 & 48.9 & +1.9 & 24.4 & 25.8 & +1.4 & 42.9 & 46.1 & +3.2 & 41.1 & 43.5 & +2.4 & 37.1 & 41.1 & +3.9 & +1.6 & +3.2 \\
\midrule
\textit{Average} & 48.7 & 54.3 & +5.6 & 58.6 & 68.0 & +9.4 & 38.1 & 52.8 & +14.7 & 51.5 & 77.3 & +25.8 & 45.2 & 67.5 & +22.3 & 46.3 & 73.9 & +27.6 & +9.9 & +25.2 \\
\bottomrule
\end{tabular}%
}
\caption{\textbf{Large oracle gaps reveal substantial headroom for frame selection.} Uniform-sampling vs.\ window-oracle accuracy (\%). Long-video gaps far exceed short-video gaps, confirming that temporal frame selection is most consequential for extended videos. Best value per column in bold.}
\label{tab:oracle}
\end{table*}

\begin{table}[t]
\centering
\resizebox{\columnwidth}{!}{%
\begin{tabular}{lrrrrr}
\toprule
Model & No Frame & Uniform & Oracle & MaxProb & MinProb \\
\midrule
InternVL3.5-8B & \textbf{38.9} & \textbf{69.1} & \textbf{76.7} & \textbf{69.5} & \textbf{67.2} \\
Qwen3-VL-8B & 37.5 & 57.2 & 66.4 & 57.6 & 56.2 \\
LLaVA-OV-7B & 36.5 & 57.0 & 65.1 & 57.0 & 54.8 \\
LLaVA-OV-0.5B & 33.0 & 46.4 & 52.1 & 47.4 & 44.4 \\
SmolVLM2-2.2B & 30.2 & 46.7 & 52.1 & 47.4 & 45.5 \\
SmolVLM2-500M & 31.2 & 42.1 & 47.7 & 42.3 & 40.7 \\
VidLLaMA3-2B & 31.9 & 39.5 & 41.0 & 38.8 & 39.1 \\
SmolVLM2-256M & 29.3 & 31.5 & 33.6 & 31.9 & 31.4 \\
\bottomrule
\end{tabular}%
}
\caption{\textbf{Window oracle dominates all strategies on MVBench.} Cross-strategy accuracy (\%) across 8 VLMs. Oracle substantially outperforms all other strategies, while MaxProb and uniform sampling perform comparably on this short-video benchmark. Best value per column in bold.}
\label{tab:cross_strategy_mvbench}
\end{table}

\begin{table}[t]
\centering
\resizebox{\columnwidth}{!}{%
\begin{tabular}{lrrrrr}
\toprule
Model & No Frame & Uniform & Oracle & MaxProb & MinProb \\
\midrule
InternVL3.5-8B & 48.8 & \textbf{80.2} & 87.7 & \textbf{80.7} & \textbf{77.0} \\
LLaVA-OV-7B & 49.2 & 78.1 & \textbf{88.2} & 78.7 & 75.0 \\
Qwen3-VL-8B & \textbf{49.8} & 70.5 & 84.4 & 71.2 & 68.9 \\
LLaVA-OV-0.5B & 35.7 & 57.0 & 69.9 & 58.5 & 54.7 \\
SmolVLM2-2.2B & 22.6 & 61.1 & 70.4 & 62.3 & 58.8 \\
SmolVLM2-500M & 25.2 & 47.2 & 57.4 & 48.3 & 45.6 \\
VidLLaMA3-2B & 30.0 & 47.0 & 48.9 & 46.4 & 46.8 \\
SmolVLM2-256M & 21.1 & 27.9 & 37.2 & 27.4 & 27.1 \\
\bottomrule
\end{tabular}%
}
\caption{\textbf{Window-oracle reveals large untapped potential on NExTQA.} Cross-strategy accuracy (\%) across 8 VLMs. MaxProb and uniform sampling remain closely matched. Best value per column in bold.}
\label{tab:cross_strategy_nextqa}
\end{table}

\begin{table}[t]
\centering
\resizebox{\columnwidth}{!}{%
\begin{tabular}{lrrrrr}
\toprule
Model & No Frame & Uniform & Oracle & MaxProb & MinProb \\
\midrule
InternVL3.5-8B & 33.0 & \textbf{64.6} & \textbf{83.8} & \textbf{66.2} & \textbf{53.2} \\
LLaVA-OV-7B & 34.6 & 63.0 & 83.4 & 62.8 & 49.0 \\
Qwen3-VL-8B & \textbf{36.6} & 54.0 & 81.0 & 58.2 & 50.2 \\
SmolVLM2-2.2B & 20.2 & 29.4 & 41.8 & 32.0 & 24.8 \\
SmolVLM2-500M & 19.4 & 26.2 & 44.4 & 26.6 & 23.2 \\
LLaVA-OV-0.5B & 15.6 & 26.4 & 40.2 & 27.4 & 21.4 \\
VidLLaMA3-2B & 25.6 & 24.4 & 25.8 & 23.0 & 23.4 \\
SmolVLM2-256M & 20.2 & 17.0 & 22.2 & 17.2 & 16.2 \\
\bottomrule
\end{tabular}%
}
\caption{\textbf{\maxprob{} leads \minprob{} on EgoSchema for 7 of 8 models.} Cross-strategy accuracy (\%) across the eight VLMs. The mean gap is $+6.5$\,pp, between the short-video and long-video groups, matching EgoSchema's three-minute clips. Best value per column in bold.}
\label{tab:cross_strategy_egoschema}
\end{table}

\begin{table}[t]
\centering
\resizebox{\columnwidth}{!}{%
\begin{tabular}{lrrrrr}
\toprule
Model & No Frame & Uniform & Oracle & MaxProb & MinProb \\
\midrule
InternVL3.5-8B & 44.9 & \textbf{65.6} & \textbf{93.7} & \textbf{72.6} & \textbf{53.4} \\
LLaVA-OV-7B & \textbf{45.1} & 63.9 & 92.7 & 70.8 & 49.9 \\
Qwen3-VL-8B & 41.4 & 52.0 & 87.3 & 60.1 & 48.6 \\
LLaVA-OV-0.5B & 35.8 & 51.3 & 82.8 & 59.9 & 43.4 \\
SmolVLM2-2.2B & 30.2 & 54.6 & 77.0 & 59.9 & 45.5 \\
SmolVLM2-500M & 31.1 & 48.6 & 78.5 & 55.2 & 41.0 \\
VidLLaMA3-2B & 34.2 & 42.9 & 46.1 & 42.8 & 41.8 \\
SmolVLM2-256M & 25.7 & 33.3 & 60.4 & 37.4 & 29.6 \\
\bottomrule
\end{tabular}%
}
\caption{\textbf{MLVU shows the strongest strategy differentiation among all benchmarks.} Cross-strategy accuracy (\%) across 8 VLMs, reflecting MLVU's high temporal demand. Best value per column in bold.}
\label{tab:cross_strategy_mlvu}
\end{table}

\begin{table}[t]
\centering
\resizebox{\columnwidth}{!}{%
\begin{tabular}{lrrrrr}
\toprule
Model & No Frame & Uniform & Oracle & MaxProb & MinProb \\
\midrule
InternVL3.5-8B & \textbf{44.6} & \textbf{57.3} & \textbf{83.9} & \textbf{61.1} & \textbf{50.3} \\
LLaVA-OV-7B & 39.7 & 56.5 & 81.8 & 59.3 & 47.9 \\
Qwen3-VL-8B & 42.8 & 48.8 & 80.9 & 52.4 & 46.1 \\
SmolVLM2-2.2B & 29.2 & 47.3 & 67.8 & 50.6 & 43.0 \\
LLaVA-OV-0.5B & 33.7 & 43.0 & 63.8 & 46.7 & 38.6 \\
SmolVLM2-500M & 31.0 & 40.9 & 69.2 & 43.8 & 38.1 \\
VidLLaMA3-2B & 36.5 & 41.1 & 43.5 & 40.9 & 40.5 \\
SmolVLM2-256M & 25.1 & 26.8 & 48.9 & 27.7 & 27.1 \\
\bottomrule
\end{tabular}%
}
\caption{\textbf{MaxProb provides meaningful gains over MinProb on LongVideoBench.} Cross-strategy accuracy (\%) across 8 VLMs, demonstrating effective frame retrieval on long videos. Best value per column in bold.}
\label{tab:cross_strategy_longvideobench}
\end{table}

\begin{table}[t]
\centering
\resizebox{\columnwidth}{!}{%
\begin{tabular}{lrrrrr}
\toprule
Model & No Frame & Uniform & Oracle & MaxProb & MinProb \\
\midrule
InternVL3.5-8B & \textbf{44.5} & \textbf{61.3} & 85.2 & \textbf{62.8} & \textbf{47.9} \\
LLaVA-OV-7B & 38.2 & 57.7 & \textbf{87.9} & 62.2 & 43.4 \\
Qwen3-VL-8B & 40.6 & 49.7 & 85.6 & 54.1 & 44.5 \\
SmolVLM2-2.2B & 28.1 & 49.4 & 73.8 & 53.2 & 39.1 \\
LLaVA-OV-0.5B & 30.7 & 44.3 & 75.0 & 49.3 & 34.7 \\
SmolVLM2-500M & 29.5 & 40.3 & 75.1 & 45.7 & 33.9 \\
SmolVLM2-256M & 25.4 & 30.8 & 67.5 & 33.7 & 27.7 \\
VidLLaMA3-2B & 28.2 & 37.1 & 41.1 & 36.8 & 37.4 \\
\bottomrule
\end{tabular}%
}
\caption{\textbf{Video-MME exhibits the largest oracle gaps among all benchmarks.} Cross-strategy accuracy (\%) across 8 VLMs. Best value per column in bold.}
\label{tab:cross_strategy_videomme}
\end{table}

\end{document}